\documentclass[letterpaper, 10 pt, conference]{ieeeconf}  % Comment this line out if you need a4paper
\usepackage[utf8]{inputenc}
\usepackage{mathtools}
\usepackage{amsmath}
\usepackage{amssymb }
\usepackage{algorithmic}
\usepackage{graphicx}
\usepackage{textcomp}
\usepackage{balance}
\usepackage{placeins}
\usepackage{stfloats}
\usepackage{float} % 提供 [H] 选项
\usepackage{mathtools} 
\usepackage{bm}
\usepackage{array}
\usepackage{longtable}
\usepackage{booktabs}
\usepackage{multirow}
\usepackage{cite}
\usepackage{csquotes}     % 增强引号处理
\DeclareMathAlphabet{\mathcal}{OMS}{cmsy}{m}{n}
\DeclareSymbolFont{largesymbols}{OMX}{cmex}{m}{n}

\IEEEoverridecommandlockouts                              
\usepackage{xcolor}
\usepackage{url}
\makeatletter
\let\NAT@parse\undefined
\makeatother
\usepackage{hyperref} 
\hypersetup{
    colorlinks=true,
    linkcolor=black,
    citecolor=green, % Set the cite color to black
    filecolor=magenta,      
    urlcolor=blue,
    }

\title{\LARGE \bf
QLIO: Quantized LiDAR-Inertial Odometry
}

\author{Boyang Lou$^{1,\diamond}$, Shenghai Yuan$^{2,\diamond}$, Jianfei Yang$^{2}$, Wenju Su$^{1}$, Yingjian Zhang$^{1}$ and  Enwen Hu$^{1,*}$,  % <-this % stops a space
\thanks{This work is supported by the National Natural Science Foundation of China under under Grants: 6220020330 and the State Key Laboratory of Heavy-duty and Express High-power Electric Locomotive Open Fund Project No.QZKFKT2024-011. }% <-this % stops a space
\thanks{$^{*}$ Corresponding Author, $^{\diamond}$ Equal Contribution. }%
\thanks{$^{1}$ Boyang Lo is with the Beijing University of Posts and Telecommunications, China. \{2022010152, suwj, gallagher, owen.hu\}@bupt.edu.cn.}%
\thanks{$^{2}$ Shenghai Yuan is with Nanyang Technological University,  Singapore, 639798. \{shyuan, jianfei.yang\}@ntu.edu.sg.}%%
}
\begin{document}

\maketitle
\thispagestyle{empty}
\pagestyle{empty}

%%%%%%%%%%%%%%%%%%%%%%%%%%%%%%%%%%%%%%%%%%%%%%%%%%%%%%%%%%%%%%%%%%%%%%%%%%%%%%%%
\begin{abstract}
LiDAR-Inertial Odometry (LIO) is widely used for autonomous navigation, but its deployment on Size, Weight, and Power (SWaP)-constrained platforms remains challenging due to the computational cost of processing dense point clouds. Conventional LIO frameworks rely on a single onboard processor, leading to computational bottlenecks and high memory demands, making real-time execution difficult on embedded systems. To address this, we propose QLIO, a multi-processor distributed quantized LIO framework that reduces computational load and bandwidth consumption while maintaining localization accuracy. QLIO introduces a quantized state estimation pipeline, where a co-processor pre-processes LiDAR measurements, compressing point-to-plane residuals before transmitting only essential features to the host processor. Additionally, an rQ-vector-based adaptive resampling strategy intelligently selects and compresses key observations, further reducing computational redundancy. Real-world evaluations demonstrate that QLIO achieves a 14.1× reduction in per-observation residual data while preserving localization accuracy. Furthermore, we release an open-source implementation to facilitate further research and real-world deployment. These results establish QLIO as an efficient and scalable solution for real-time autonomous systems operating under computational and bandwidth constraints.
%As edge robotic platforms equipped with LiDAR sensors and inertial measurement units (IMU) become prevalent, enabling efficient spatial perception under size, weight and power (SWAP) constraints remain challenging. However, the substantial bandwidth consumption of raw point clouds across multiple processing units for state estimation severely limits deployment on resource-constrained devices. To address this, we propose the first quantized LiDAR-inertial odometry (QLIO) that dramatically reduces data transfer while maintaining estimation accuracy. We introduce two key innovations: (1) A multi-processor distributed state estimation scheme that quantizes point-to-plane associations and utilizes QMAP for efficient estimation. (2) An rQ-vector-based adaptive resampling strategy that selectively downsamples and compresses key measurements to minimize bandwidth usage. Extensive evaluations demonstrate that our the  method reduces per-observation residual data by 7× through quantization, and achieves remarkable 200x bandwidth reduction compared to direct cloud compressed. Notably, QLIO maintains comparable accuracy to conventional approaches despite aggressive residuals quantization, making it particularly suitable for SWAP-constrained edge robotics.

\end{abstract}

%\textbf{\textit{Index Terms} --- Blind Spot Detection, Autonomous Vehicles, Advanced Driver-Assistance Systems }
\section*{Supplementary Material}
The video, code, and supplementary materials will be available at:  \url{https://github.com/luobodan/QLIO}

%%%%%%%%%%%%%%%%%%%%%%%%%%%%%%%%%%%%%%%%%%%%%%%%%%%%%%%%%%%%%%%%%%%%%%%%%%%%%%%%

\section{Introduction}

The rise of the low-altitude economy \cite{yuan2024mmaud} and humanoid robotics \cite{he2024omnih2o,li2024hcto,li2025helmetposer} is accelerating demand for efficient, accurate, and depth-aware 3D navigation. As these systems expand into general outdoor dynamic environments, real-time localization and mapping must be adapted to meet increasing constraints on computational efficiency \cite{xu2022fast,bai2022faster,nguyen2024eigen}, power consumption, and deployment feasibility.

LiDAR-Inertial Odometry (LIO) and SLAM have been widely explored for robotic perception and navigation, with numerous solutions \cite{shan2020lio,nguyen2021miliom,shan2021lvi,xu2022fast,bai2022faster,lim2023adalio,nguyen2023slict,vizzo2023kiss,chen2023direct,chen2024ig,ji2024lio,wu2024lio,nguyen2024eigen} improving mapping accuracy \cite{eisoldt2022fully,lin2022r,jin2024robust,Li2024PSS,Li2024graph} and localization robustness \cite{ma2024mm,yin2024outram,Zhao2024adaptive}. However, existing methods remain \textbf{heavily biased toward} algorithmic performance metrics \cite{rebecq2016evo,wang2020tartanair,helmberger2022hilti,nair2024hilti} such as Absolute Pose Error (APE) and Relative Pose Error (RPE), often neglecting feasibility for lightweight, real-time applications \cite{wang2017non,yang2024fast}. While some approaches improve runtime efficiency to operate on Size, Weight, and Power (SWaP)-constrained platforms \cite{xu2022fast,bai2022faster,nguyen2024eigen,zheng2024fast}, they often sacrifice robustness. 
%LiDAR-Inertial Odometry (LIO) and SLAM have been extensively studied, with numerous solutions improving mapping accuracy \cite{eisoldt2022fully,lin2022r,jin2024robust,Li2024PSS,Li2024graph} and localization robustness \cite{ma2024mm,yin2024outram,Zhao2024adaptive}. However, most methods prioritize algorithmic performance metrics \cite{rebecq2016evo,wang2020tartanair,helmberger2022hilti,nair2024hilti} such as Absolute Pose Error (APE) and Relative Pose Error (RPE), often at the expense of real-time feasibility \cite{wang2017non,yang2024fast}. While some approaches improve runtime efficiency for SWaP-constrained platforms \cite{xu2022fast,bai2022faster,nguyen2024eigen,zheng2024fast}, they frequently compromise robustness.

\begin{figure}
\centering
  \includegraphics[width=1\linewidth]{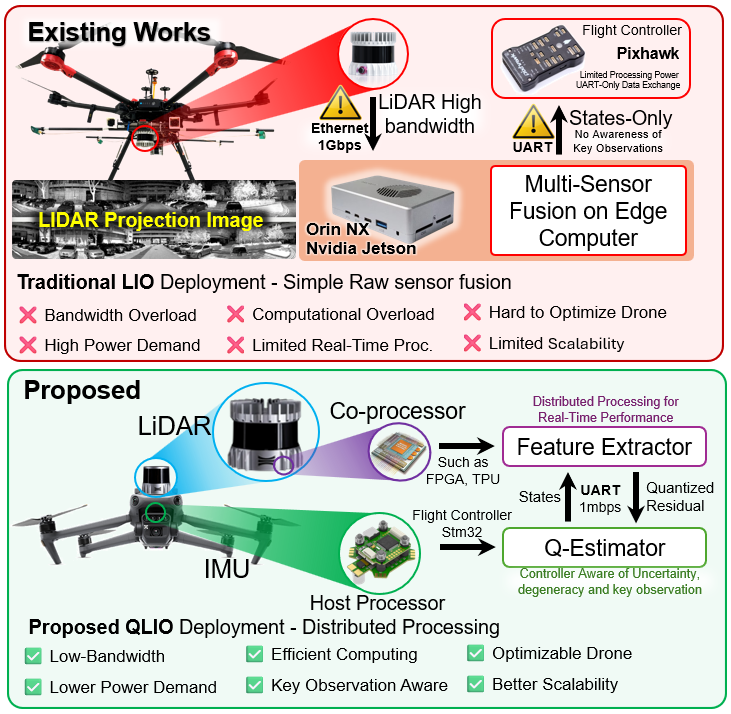}
  \vspace{-22pt}
  \caption{Comparison of Traditional LIO deployment \cite{nguyen2021viral, nguyen2022ntu,nguyen2024mcd} vs. Proposed QLIO: Enhancing Efficiency with Distributed Processing.}
  \label{motivation}
    \vspace{-22pt}
\end{figure}

While integrating visual perception \cite{lin2022r,xu2023airvo,pfreundschuh2024coin,zheng2024fast,xu2025airslam} improves state estimation, it increases computational complexity. The cloud-based LiDAR localization \cite{ribeiro2006soi,trawny2009cooperative,msechu2008decentralized} offloads processing but suffers from latency and network constraints. Efficient compression schemes \cite{schnabel2006octree,draco_github,graziosi2020overview,wang2022r,wang2022sparse} help address bandwidth limitations, yet balancing compression efficiency and data fidelity remains challenging. Meanwhile, advances in SoC architectures \cite{jiang2024live} and distributed processing frameworks \cite{peng2024quantized} are shifting the research focus to edge computing, optimizing real-time compression, data transmission, and localized processing under strict SWaP constraints.
%While visual perception integration \cite{lin2022r,xu2023airvo,pfreundschuh2024coin,zheng2024fast,xu2025airslam} enhances state estimation, it introduces optimization challenges and computational overhead. Similarly, cloud-based LiDAR localization \cite{ribeiro2006soi,trawny2009cooperative,msechu2008decentralized} reduces onboard processing but suffers from latency and network constraints, limiting real-time feasibility. To address bandwidth limitations, efficient compression schemes \cite{schnabel2006octree,draco_github,graziosi2020overview,wang2022r,wang2022sparse} have been explored, yet balancing compression efficiency and data fidelity remains a challenge. Meanwhile, advancements in SoC architectures for 2D LiDARs \cite{jiang2024live} and distributed processing frameworks \cite{peng2024quantized} have enabled on-device computation, reducing cloud dependency. These trends indicate a shift toward edge computing, where real-time compression, efficient data transmission, and localized processing must be optimized to achieve scalability under strict SWaP constraints.

SoC and distributed processing \cite{jiang2024live, peng2024quantized,huai2024consistent}  have proven effective for 2D visual pipelines. However, deploying such a solution to 3D LIO \cite{cai2025bev,li2024ua,li2025limo} in real-world systems remains challenging due to the trade-off between SWaP and robustness. Robust algorithms require substantial processing power \cite{ji2024sgba}, yet practical platforms, such as UAVs and humanoids, must remain lightweight and efficient. This trade-off limits scalability, demanding a rethink of LIO architectures for adaptive processing and efficiency.
%The Soc or distributed processing has been proven working for visual pipline.However, deploying LiDAR-Inertial Odometry (LIO) in real-world systems still faces a \textbf{challenging} trade-off between computational demand and system limitations. High-accuracy algorithms require substantial processing power \cite{ji2024sgba}, yet practical platforms—from UAVs to humanoid robots—must remain lightweight and power-efficient. This trade-off forces compromises between accuracy, robustness, and real-time performance, hindering scalable deployment. Addressing these challenges demands a rethinking of LIO architectures, focusing on adaptive processing, computational efficiency, and scalable solutions for autonomous systems.

To bridge this gap, we propose QLIO, a multi-processor distributed quantized LIO framework tailored for edge devices. Unlike prior work such as Quantized Visual-Inertial Odometry (QVIO) \cite{peng2024quantized}, which applies quantization-based estimation to visual features, QLIO introduces LiDAR-specific optimizations that enhance efficiency while maintaining localization accuracy. Our method employs structured point-to-plane residual compression and rQ-vector-based adaptive resampling to reduce computational load and bandwidth consumption without compromising geometric integrity. Our contribution can be summarized as follows.

% \begin{itemize}
%     \item Multi-Processor Distributed LIO:
% We introduce the first co-processor and host-processor distributed architecture for LIO, where quantized point-to-plane residuals are preprocessed on the co-processor and transmitted efficiently to the host for state estimation, reducing computational and bandwidth overhead.
%     \item Quantized Maximum A Posteriori Estimation (QMAP):
% We propose a multi-bit quantized MAP estimation framework, integrating quantized residuals into the optimization process. Unlike existing binary quantization methods, our multi-bit adaptive quantization balances compression and accuracy, making LIO scalable for real-time applications.
%     \item Adaptive Resampling and Compression with rQ-Vectors:
% We introduce an rQ-vector-based adaptive resampling method that selects informative measurements and groups quantized residuals, achieving efficient compression while maintaining key geometric information for accurate localization.
%     \item Extreme Bandwidth Reduction with Real-World Validation:
% QLIO achieves a 7× per-point bandwidth reduction (down to 3.7 bytes per observation) and a 200× overall bandwidth reduction compared to traditional point cloud compression. We validate QLIO’s scalability and efficiency across real-world datasets, demonstrating its feasibility for autonomous systems and edge computing applications.
% \end{itemize}

\begin{figure*}[t]
    \centering
    \includegraphics[width=0.9\linewidth]{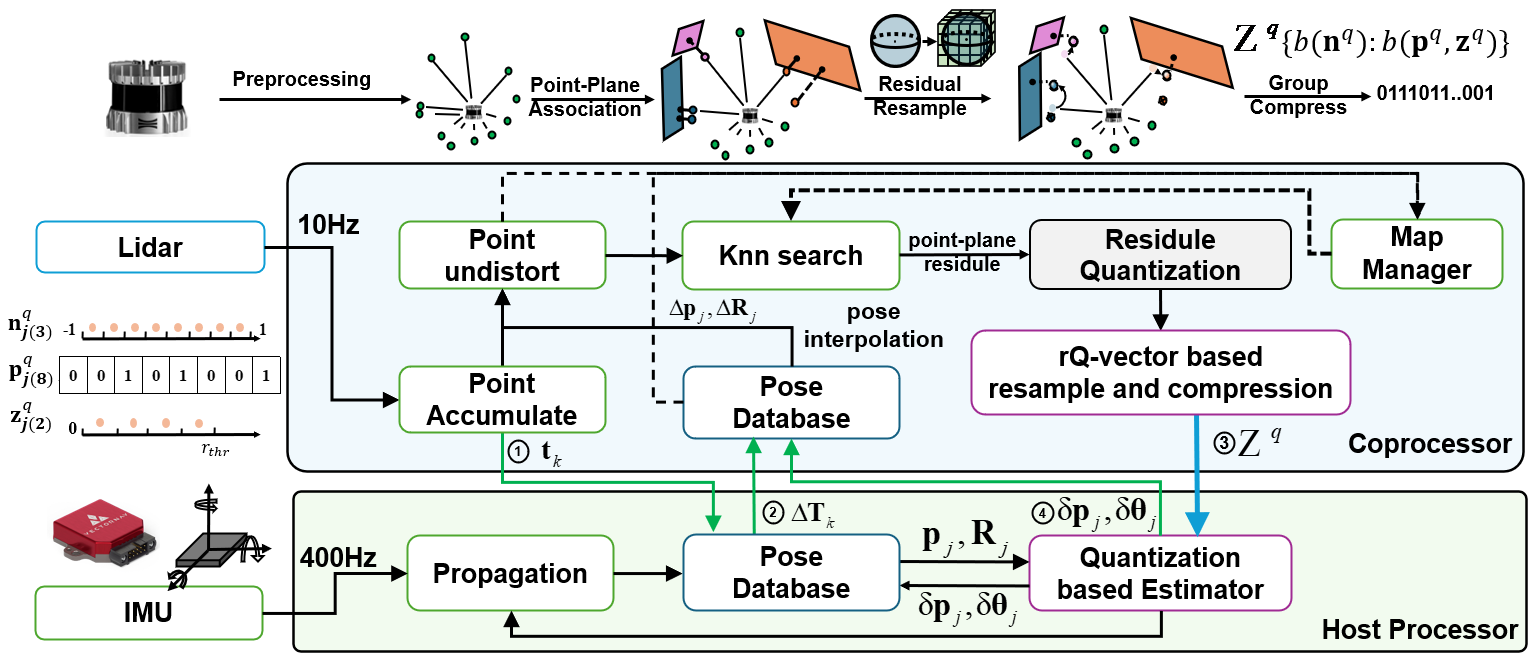} 
    \vspace{-15pt}
    \caption{QLIO system architecture with a dual-processor setup. The coprocessor handles LiDAR preprocessing, residual quantization, and compression, while the host processor performs IMU propagation and quantized state estimation, optimizing efficiency for real-time localization.}
    \label{fig2}
     \vspace{-10pt}
\end{figure*}

\begin{itemize}
\item Distributed Multi-Processor LIO Architecture:
We propose QLIO, a distributed LiDAR-Inertial Odometry framework, leveraging a co-processor for LiDAR preprocessing and a host processor for state estimation. This reduces computational load and bandwidth usage, specifically tailored for LiDAR data.
\item Quantized MAP Estimation for LIO:
We introduce a multi-bit quantized Maximum A Posteriori (QMAP) estimator for LIO, optimizing 3D LiDAR residuals, reducing data transmission while preserving localization accuracy.
\item  rQ-Vector-Based Adaptive Resampling and Compression:
We present an rQ-vector-based resampling and compression technique, focusing on LiDAR's 3D geometric structure to improve data efficiency and minimize computational redundancy.

\item Open-Source LIO Framework with Real-World Validation:
Our open-source QLIO framework is validated on real-world datasets, demonstrating significant bandwidth reduction (14.1×) while maintaining high localization accuracy, making it ideal for SWaP-constrained systems.
\end{itemize}

\section{Related Works}
% As there is not some quantized QLIO algorithm exists.
%\subsection{Data quantization in LiDAR slam}\label{AA}
% \subsection{Soc accelerate of the LiDAR slam}
To reduce the computational load of the state estimation, LiDAR-SLAM systems \cite{cao2025cooperative,liang2025unsupervised,xu2024selective,deng2024compact,Nguyen2024GPTR,bai2024collaborative,yu2024i2ekf,Li2024PSS,li2024hcto,yin2024outram,nguyen2024eigen,ji2022robust,nguyen2021liro} often utilize the structural 
characteristics to quantize the point cloud measurements through feature extraction\cite{zhang2014loam,shan2018lego} 
or voxel downsampling\cite{xu2022fast,bai2022faster}. In subsequent work, downsampling based on an octree structure was proposed \cite{liu2024voxel}, 
as well as adaptive downsampling methods \cite{lim2023adalio} to improve performance in degradation scenarios. Y.C et al. proposed a 
loop-closure BTC descriptor \cite{yuan2024btc}, 
which obtains a lightweight representation by using binary height description. 
Quantized state estimation, though underexplored, has a foundation in observation quantization. The field originated from the SOI-KF framework~\cite{ribeiro2006soi}, which introduced 1-bit residual sign discrimination to balance information retention and efficiency. Later advancements extended quantization theory to distributed architectures through multi-bit MAP estimation and IQEKF variants~\cite{nerurkar2013communication, msechu2008decentralized, trawny2009cooperative}. QVIO~\cite{peng2024quantized} further developed this by integrating dual quantization strategies: zQVIO for multi-bit observation quantization and rQVIO for 1-bit residual quantization, achieving millimeter precision with minimal bandwidth. While effective for VIO, QVIO’s 2D pixel-based quantization does not generalize to LIO, which depends on 3D point-cloud geometry. Addressing this gap, QLIO employs structured point-to-plane residual quantization and adaptive rQ-vector grouping for efficient LiDAR data compression while preserving geometric consistency. This innovation bridges quantization theory to LIO, ensuring accurate and bandwidth-efficient processing.

% The field evolved from Nerurkar et al.'s seminal SOI-KF framework~\cite{ribeiro2006soi}, 
% which pioneered observation quantization through 1-bit residual sign discrimination.
% This approach leveraged Gaussian posterior assumptions and tail probability analysis 
% to minimize information loss while enabling efficient state estimation.
%  Subsequent developments introduced multi-bit maximum a posteriori estimation 
%  and IQEKF variants~\cite{nerurkar2013communication,
%  msechu2008decentralized,trawny2009cooperative}, 
%  systematically extending quantization theory to distributed processing architectures.
%  Building upon these foundations, Y.P et al. introduced QVIO~\cite{peng2024quantized} 
% ——a visual-inertial quantized estimator employing dual quantization strategies: 
% zQVIO implements multi-bit observation quantization while rQVIO applies 1-bit residual quantization. 
% Through rigorous noise propagation modeling and Gaussian tail probability optimization, 
% both variants achieve millimeter-level precision with sub-100bps bandwidth consumption, 
% demonstrating the framework's effectiveness in resource-constrained scenarios.
% In the subsequent QVIO2, it was extended by using differential quantization and multi - bit 
% quantization for observations and residuals, and a unified tail probability interval state estimation method
%  was introduced in the backend to estimate the overall state. Compared with the its baseline-Openvins\cite{geneva2020openvins}, it achieved 
%  improved accuracy while maintaining a low number of observations/bits.

%  width=0.5\textwidth

\section{Quantized LiDAR-Inertial-Odometry}
Our framework inherits architectural principles from FastLIO2~\cite{xu2022fast} and QVIO~\cite{peng2024quantized}, assuming fixed and pre-calibrated LiDAR-IMU extrinsic 
parameters excluded from the optimization manifold. The system maintains the core state vector defined 
in FastLIO2:
 \begin{gather}
    % \begin{split}
\mathbf{{x}} \triangleq \left[ 
    % \{G}\mathbf{\mathit{R}}_{I}^{T}, 
    {\mathop{^{G} {\mathbf{R}}_{I}^{T}}},
    {\mathop{^{G} {\mathbf{p}}_{I}^{T}}},
    {\mathop{^{G} {\mathbf{v}}_{I}^{T}}},
    \mathbf{\mathit{\mathbf{b}}}_{\omega}^{T}, 
    \mathbf{\mathit{\mathbf{b}}}_{a}^{T}, 
    {\mathop{^{G} {\mathbf{g}}_{I}^{T}}}
 \right]^{T} \in \mathcal { M }
\\
\mathbf { u } \triangleq \left[ \omega _ { m } ^ { T } 
\quad \mathbf { a } _ { m } ^ { T } \right] ^ { T } , 
\mathbf { w } \triangleq \left[ \mathbf { n } _ 
{ \omega } ^ { T } \quad \mathbf { n } _ { \mathrm { a } }
 ^ { T } \quad \mathbf { n } _ { \mathrm { b } \omega } ^
  { T } \quad \mathbf { n } _ { \mathrm { b a } } ^ { T } \right] ^ { T }
  \\
    \mathbf { x } _ { i + 1 } = \mathbf { x } _ { i } \boxplus  
    \left( \Delta t \mathbf { f } \left( \mathbf { x } _ { i } , 
    \mathbf { u } _ { i } , \mathbf { w } _ { i } \right) \right) 
% \end{split}
 \end{gather}\label{eqkine}
 $\mathbf{R}$, $\mathbf{p}$, and $\mathbf{v}$ represent rotation, translation, and velocity, respectively. $\mathit{\mathbf{b}}_{a}$ 
 and ${\mathbf{b}}_{\omega}$ are acceleration biases and the angular velocity of the IMU.
  $\mathbf{g}$ denote the gravity and  $\mathbf { w }$ and $\mathbf { u }$ 
  denote IMU measurement noise and control input. 
  \textbf{Global frame} $^G(\text{·})$ represents the World-fixed reference frame initialized at system startup. \textbf{Body frame} $^I(\text{·})$ rigidly attached to IMU with origin at its measurement center, establishing the inertial reference for kinematic computations and \textbf{LiDAR frame} $^L(\text{·})$ defined by LiDAR’s optical center, where raw point clouds are initially captured in this right-handed Cartesian system.
\subsection{System Overivew}\label{BB}
% Our system comprises a host-processor (responsible for integrating the IMU and estimating the state) 
% and a custom-designed co-processor (responsible for point cloud processing, local point cloud nearest
%  neighbor search, and residual calculation) as shown in figure \ref{fig2}. 
%  When the IMU data arrives at the host-processor, 
%  it is integrated using the results obtained from the previous optimization, 
%  and the integrated pose is saved to the pose database.
% The points will be accumulated to form a complete scan and record the every arrived time. 
%   The two processors always maintain a complete communication link and synchronized timing. 
%   The processment include 4 steps just like in~\cite{peng2024quantized}:\\
Our system architecture comprises two synchronized components as illustrated in \autoref{fig2}. 
% a host-processor handling IMU data integration and state estimation, 
% coupled with a custom co-processor dedicated to point cloud processing 
% tasks including nearest neighbor search and residual calculation, 
% as illustrated in Figure \ref{fig2}. 
  \begin{equation}
    \begin{split}
        \mathbf{x}_{i+1}=\mathbf{x}_{i} \boxplus \left( \Delta t \mathbf { f } \left( \widehat { \mathbf { x } } _ { i } ,
         \mathbf { u } _ { i } , \mathbf { 0 } \right) \right) ,
        \widehat { \mathbf { x } } _ { 0 } = \overline { \mathbf { x } } _ { k - 1 } ,
        \\
        \widehat { \mathbf { P } } _ { i + 1 } = \mathbf { F } _ { \mathbf { x } _ { i } } 
        \widehat { \mathbf { P } } _ { i } \mathbf { F } _ { \mathbf { x } _ { i } } ^ { T } 
        + \mathbf { F } _ { \mathbf { w } _ { i } } \mathbf { Q } _ { i } \mathbf { F } _ 
        { \mathbf { w } _ { i } } ^ { T } ; \widehat { \mathbf { P } } _ { 0 } = 
        \overline { \mathbf { P } } _ { k - 1 }
    \end{split}\label{forward}
\end{equation}
% The host-processor perform
% integration and propagation (\autoref{forward}) upon IMU data arrival, 
% recording the timestamp and maintaining a continuously updated to the pose database. 
% The co-processor dedicated to point cloud processing 
% tasks including nearest neighbor search and residual calculation.
% The architecture ensures tight inter-processor synchronization    through maintained communication links and hardware-level clock
%     alignment, following a four-stage processing pipeline inspired by prior works~\cite{peng2024quantized}:
The host processor performs IMU integration and pose updates (\autoref{forward}), while a dedicated co-processor efficiently handles point cloud processing, including nearest neighbor search and residual computation. Unlike Visual-Inertial Odometry (VIO) systems that rely on image-based feature tracking and quantized pixel measurements~\cite{peng2024quantized}, our method is tailored for LiDAR-Inertial Odometry (LIO), leveraging structured point-to-plane residual quantization to compress geometric information efficiently. By integrating a quantization-aware four-stage pipeline, our architecture minimizes bandwidth and computational costs while maintaining precise inter-processor synchronization.
 \\1. Upon LiDAR scan arrival at timestamp $\mathbf{t}_k$, 
 coprocessor sent the pose request with the timestamp 
 $\mathbf{t}_k$ to the host-processor.\\
 2. The host-processor responds with the relative pose transform 
 $\Delta\mathbf{T}_k =\left( \Delta \mathbf{p}_k, \Delta {\mathbf{R}}_{k} \right) $ 
 obtained by the short term IMU integration from $\mathbf{t}_{k-1}$ to $\mathbf{t}_k$. 
 Then the coprocessor downsample and undistorted the points
 using the relative pose interpolation and transform points to the global coordinate system:
\begin{equation}  
    \begin{split}
        ^{L}\mathbf{p}_{j} &={^{L}\mathbf{T}_{I}} \Delta\mathbf{T}_{j}{^{I}\mathbf{T}_{L}}\mathbf{p}_{j}
        \\
        ^{G}\mathbf{p}_{j} &= \Delta\mathbf{T}_k\mathbf{T}_{k-1} {^{I}\mathbf{T}_{L}}{^{L}\mathbf{p}_{j}}
    \end{split}
\end{equation} 
where $\Delta\mathbf{T}_{j} =\left( \Delta \mathbf{p}_{j}, \Delta {\mathbf{p}_{j}}_{ori} \right) $ respected to 
the LiDAR relative pose from the ${{\mathbf{p}}_{j}}_{ori}$ arrived time $\mathbf{t}_j$ 
to the end of the scan $\mathbf{t}_k$.
Then using the $^{G}\mathbf{p}_{j}$ perform knn search based on map manager, finding the nearest five points. If the 
they can fit an plane, we will find the associations between points and plane, as well as the residual measurements $\mathbf{z}_{i}$
   in \cite{xu2022fast}:%$\mathbf{\mathrm{n}}$, $\mathbf{\mathrm{p}}$
%  \begin{equation}
%     \begin{aligned}
%     \mathbf{z}_{i} = \mathbf{u}_{i}^{T} 
%     \left( ^{G}\mathbf{p}_{\mathbf{i}} - 
%     {}^{G}\mathbf{m}_{i} \right) = d_{i} \cdot \mathbf{n}_{\mathbf{i}} 
%     \\begin{align*}
%         \left\| \mathbf{n}_{\mathbf{i}} \right\| &= 1 , \\
%         \left\| \mathbf{z}_{i} \right\| &= d_{i}
%     \end{align*}
%     \end{aligned}
% \end{equation}\frac
% \begin{equation}
    \begin{gather}
        \mathbf{z}_{i} = \mathbf{u}_{i}^{T} \left( {}^{G}\mathbf{p}_{i} - {}^{G}\mathbf{m}_{i} \right) = \left\| \mathbf{n}_{i} \right\|, \quad 
        \mathbf{n}_{i} = d_{i} \cdot \frac{\mathbf{u}_i^T}{\left\|\mathbf{u}_i^T\right\|} \\
        0 = \mathbf{h}_{i}\left(\mathbf{x}_{k}, {}^{L}\mathbf{p}_{i} + {}^{L}\mathbf{n}_{i}\right) \\
        \mathbf{z}_{i} + \mathbf{h}_{i}(\mathbf{x}_{k} - \hat{\mathbf{x}}_{k}) = -\mathbf{v}_{i} \sim \mathcal{N}\left(0, \mathbf{R}_{i}\right) \label{measurment eq}
    \end{gather}
% \end{equation}
where ${\mathbf{v}_{i}}$ and ${\mathbf{R}_{i}}$ denote the 
measurment noise and covariance matrix.
Residual vector {$\mathbf{n}$} is obtained by
 projecting the displacement vector ({$\mathbf{p}_{i}-\mathbf{m}_{i}$}) 
 onto the normal vector {$\mathbf{u}_i^T$}
  of the associated plane.
\\3.Once point-plane association is established, the overall state estimation problem 
will solely depend on the associated points and residual vectors. 
 We can use $s$ bits quantize them with
$\mathbf{f}_{\mathbf{\mathit{p}}_{(s)}}^{\mathbf{\mathit{q}}} $,
$\mathbf{f}_{\mathbf{\mathit{z}}_{(s)}}^{\mathbf{\mathit{q}}} $,
$\mathbf{f}_{\mathbf{\mathit{n}}_{(s)}}^{\mathbf{\mathit{q}}} $
and perform resampling(Section \ref{3sectionB}):
\begin{equation}
% \begin{gather*}
    \label{z_eq}
    \begin{aligned}
    \mathbf{z}_{i}^{q} &= \mathbf{b} \left(\mathbf{z}_{i}^{}\right) = 
    \mathbf{f}_{\mathbf{\mathit{z}}{(s)}}^{\mathbf{\mathit{q}}}
     \left(\mathbf{z}_{i}^{}\right) \\
    % \mathop{^{G} {R}_{I}^{T}} 
    {{^{L}}}\mathbf{p}_{\mathbf{\mathit{i}}}^{\mathbf{\mathit{q}}} &= 
    \mathbf{f}_{\mathbf{\mathit{p}}{(s)}}^{\mathbf{\mathit{q}}} \left(
        ^{L}\mathbf{p}_{\mathbf{\mathit{i}}} \right),
    {{^{G}}}\mathbf{n}_{\mathbf{\mathit{i}}}^{\mathbf{\mathit{q}}} = 
    \mathbf{f}_{\mathbf{\mathit{n}}{(s)}}^{\mathbf{\mathit{q}}} \left(
        ^{G}\mathbf{n}_{\mathbf{\mathit{i}}} \right)     
    \end{aligned}
% \end{gather*}
\end{equation}
and then zip the all quantization result
${\mathbf{z}^{q},\mathbf{p}^{\mathbf{\mathit{q}}}, \mathbf{n}^{\mathbf{\mathit{q}}}}$ 
to a group, $\mathcal{Z}^q$, and compressed to the bitstream, transmitting them to the host processor.
\\4.The host processor performs a quantization-based estimator (Section \ref{3sectionC}) and updates the pose to the database.

\begin{figure}[t]
    \centering
    \includegraphics[width=1\linewidth]{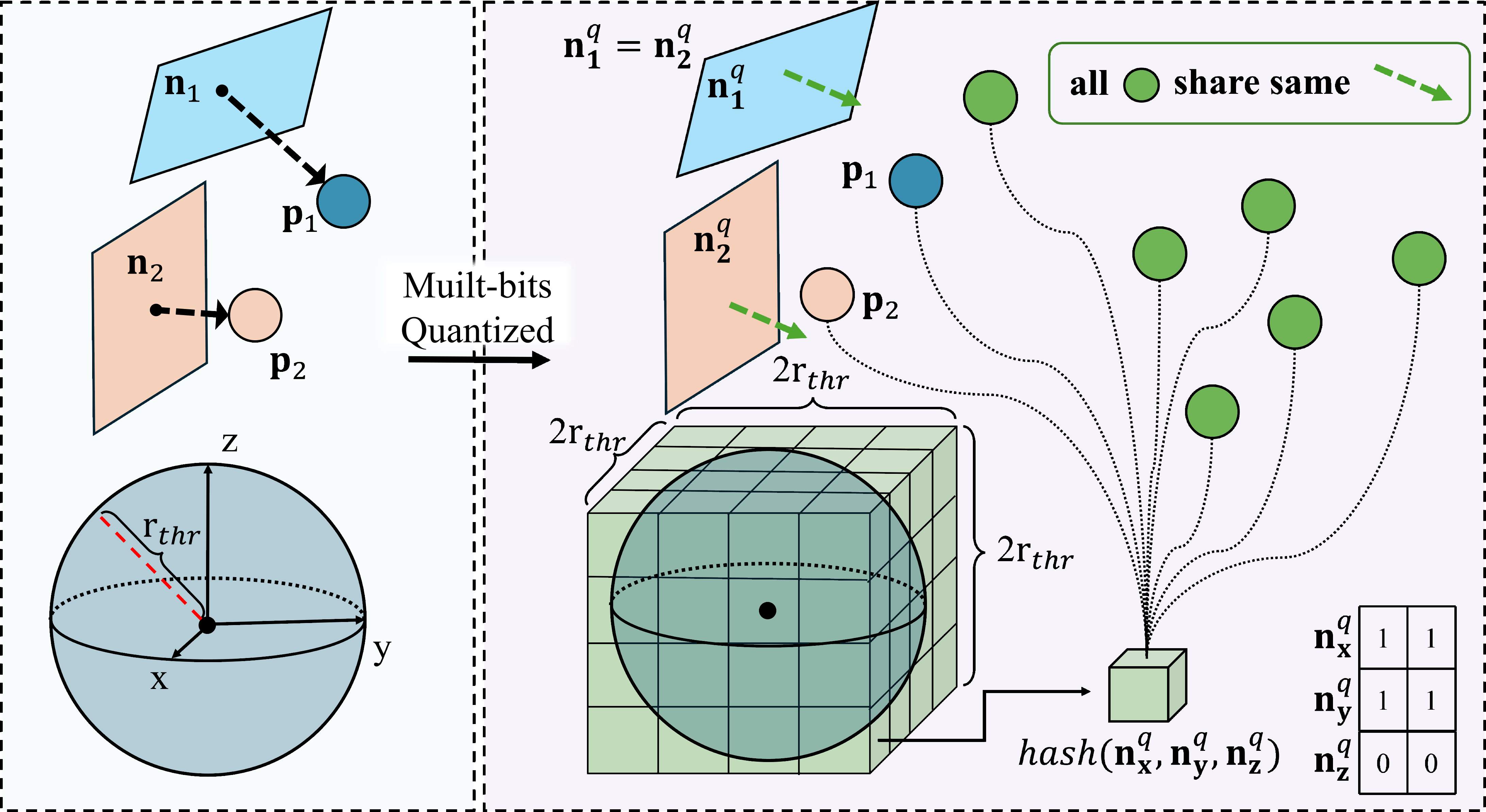}
    \vspace{-22pt}
    \caption{The residual vector space is quantized into a voxel space based on INT-2 quantization. 
    3D points with the same quantized residuals are grouped into the same hash index. 
    Additionally, the point clouds can be compressed together according to the hash index.}
    \label{fig_rqvector}   
     \vspace{-15pt}
\end{figure}

\subsection{rQ-vector based Resampling and Group Compression}\label{3sectionB}\
Although point cloud observations have undergone preliminary filtering 
through voxel downsampling and point-plane association, 
the remaining points and residual vectors (28 bytes per measurement based on Float32) remain computationally intensive for direct transmission, 
consuming substantial communication bandwidth. 
% Moreover, the residual observations may exhibit significant redundancy 
% and directional imbalance, which may \textit{degrade} positioning accuracy.
As shown in figure~\ref{fig_rqvector}, we propose a quantization-based resampling 
approach that leverages 
quantized residual vector to construct hash indices and group the points,
then adaptive downsample the points those shared with same rQ-vector.
% This method enables directional downsampling of observations 
% and their associated residuals, yielding a more uniform spatial distribution
% with reduced quantity. The directional quantization process creates \(2^m\) 
% spatial partitions per dimension (where \(m\) denotes the quantization bit-width), 
% effectively implementing \textit{normal vector voxelization} as illustrated in Figure~\ref{fig}. 
If we assume residual $\mathbf{z}=\left\|\mathbf{n}_{i} \right\|<{r}_{thr}$ ,
the quantization discretizes residual vectors by allocating $s$ bits per axis, creating $2^s$ intervals 
in each dimension that voxelize the normal vector space. This produces $2^{3s}$ 
unique space cells $\mathcal{C}_k$ where $k \in [1,2^{3s}]$, 
each corresponding to a specific rQ-vector range. 
For any observed point $\mathbf{p}_i$ 
with original residual
 vector $\mathbf{n}_i$, the quantization process
 $\mathbf{f}_{\mathbf{\mathit{p}}(s)}^{\mathbf{\mathit{q}}} $
 will generate a 3$s$-bit hash key. All points sharing identical $\mathbf{n}^q$ 
 are aggregated into the same hash bucket $\mathcal{H}_k$, 
 where the bucket index $k$ can computed by hash function:
\begin{equation}
\mathbf{p}_{\mathbf{\mathit{i}}}^{\mathbf{\mathit{q}}} \in 
\mathcal{H}_k, k=hash(\mathbf{n}_{\mathbf{\mathit{i}}}^{\mathbf{\mathit{q}}})
\end{equation}
This spatial hashing ensures that the points 
with similar residuals are clustered into identical buckets. We then perform
 an adaptive 3D downsampling process to each buckets, where downsample
 size ${ds}_k$ can calculate as follows:
 \begin{equation}
    {ds}_k = {ds}_{0} + \alpha \sum_{i=1}^{n_k} \frac{1}{n_k}\left\|{\mathbf{\mathbf{p}}_i} \right\|
\end{equation}
where ${ds}_{0}$ is the cloud preprocessing downsampling parameter and 
$\alpha$ is the distance penalty coefficient derived from the average 
Euclidean distance of the points within the same bucket. 
 
Additionally, the filtered points and their corresponding 
residuals can subsequently be grouped based on rQ-vectors, 
employing a group encoding scheme with shared header structures 
to reduce transmission bandwidth. And the groups $\mathcal{Z}^q$ can 
then be constructed as:
\begin{equation}
    \begin{split}
    \mathcal{Z}^q &= 
    \left\{b(^{1}\mathbf{n}^{\mathbf{\mathit{q}}}_{k}):
    b({^{1}\mathbf{z}^{q}_{k}},
    {^{1}\mathbf{p}^{\mathbf{\mathit{q}}}_{k}}),
    b({^{2}\mathbf{z}^{q}_{k}},
    {^{2}\mathbf{p}^{\mathbf{\mathit{q}}}_{k}}),
    b({^{3}\mathbf{z}^{q}_{k}},
    {^{3}\mathbf{p}^{\mathbf{\mathit{q}}}_{k}}),... \right\},
    \\
    &\left\{b({^{2}\mathbf{n}^{\mathbf{\mathit{q}}}_{k+1}}):
    b(^{1}{\mathbf{z}^{q}_{k+1}},
    {^{1}\mathbf{p}^{\mathbf{\mathit{q}}}_{k+1}}),
    b({^{2}\mathbf{z}^{q}_{k+1}},
    {^{2}\mathbf{p}^{\mathbf{\mathit{q}}}_{k+1}}),... \right\}...
    \end{split}\label{compressed_stragy}
\end{equation}
\vspace{-5pt}
where $b(\text{·})$ denotes the multi-bit binary quantization.

  \begin{figure}
\centering
  \includegraphics[width=0.9\linewidth]{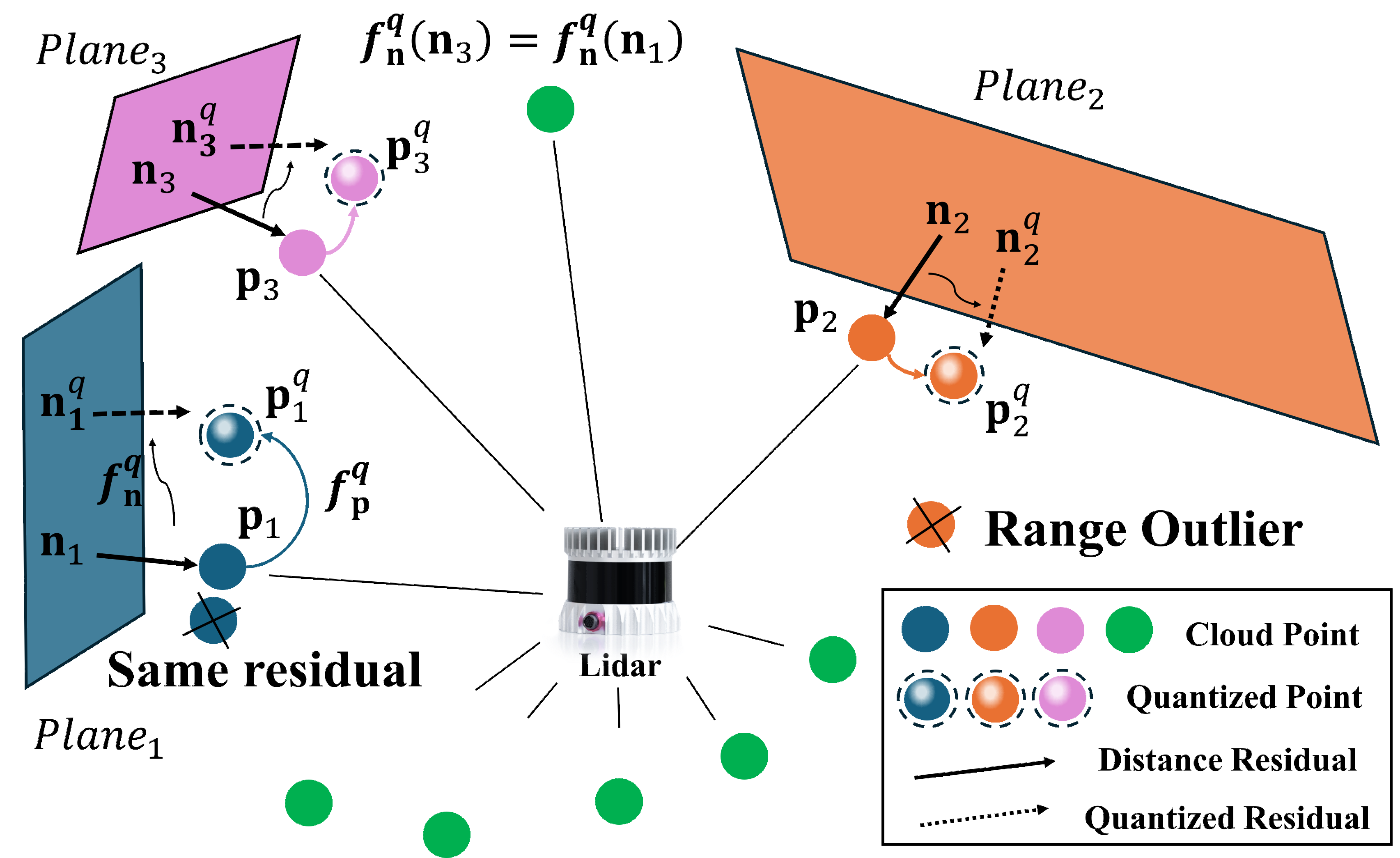}
  \vspace{-12pt}  
  \caption{Quantization of the points and their association point-plane residuals. }
  \label{processing}
  \vspace{-20pt}
\end{figure}

\subsection{Quantization-based Estimation}\label{3sectionC}\
% After the host processor receives the quantization measurements, 
% it solves the overall optimization problem by referring 
% to the state estimation formulas in [fastlio][IBQKF][qvio]. 
The quantization-based estimation problem can formulate as a MAP problem 
\cite{trawny2009cooperative}\cite{peng2024quantized}:
\begin{equation}
    \begin{aligned}
 \arg \max \: &p\left(\mathbf{x} | \mathbf{\mathbf{b}} \left(\mathbf{z}\right)\right) \\
% =p\left(\mathbf{x} | \mathbf{b}(\mathbf{z}) \right\)\\ % = \left[b\left(\mathbf{z}_{1}\right),b\left(\mathbf{z}_{2}\right),...,b\left(\mathbf{z}_{i}\right)\right] \right)
= &p \left(\mathbf{x}\right)  
 \prod_{i = 1}^{m} 
 p \left(\mathbf{\mathbf{b}} \left(\mathbf{z}_{i}\right)|\mathbf{\mathbf{x}}_{i}\right)
 \\
 = &p \left(\mathbf{x}\right) 
 \prod_{i = 1}^{m} 
 p\left(\mathbf{\mathbf{b}}(\mathbf{z}_{i})\in\left[\mathbf{z}^{0}_{i},\mathbf{z}^{1}_{i}\right]
|\mathbf{\mathbf{x}}_{i}\right)
    \end{aligned}\label{MAP_problem}
\end{equation}
The original measurements of the quantized observation residuals {$\mathbf{z}^q$}
should fall within the quantization interval between {$\mathbf{z}^0$} and {$\mathbf{z}^1$}. Using 
\autoref{measurment eq} and  $Q$ fuction
to :
\begin{equation}
    \begin{split}
        p\left(\mathbf{\mathbf{b}}(\mathbf{z})|\mathbf{x}\right) 
        &= \mathbf{Pr}\left\{\mathbf{z}^{0} <-h\left(\mathbf{x} - \hat{\mathbf{x}}\right) - 
        \mathbf{v}<\mathbf{z}^{1} \middle| \mathbf{x} \right\}
\\
&=\mathbf{Pr} \left\{ \frac{\mathbf{z}^{0}}{\sigma} < 
\frac{- h \left( \mathbf{x} - \hat{\mathbf{x}} 
\right) - \mathbf{v}}{\sigma} < \frac{\mathbf{z}^{1}}{\sigma} \, 
\middle| \, \mathbf{x} \right\} 
\\
&= Q \left( \mathbf{\mathcal{X}}^{1} \right) 
- Q \left( \mathbf{\mathcal{X}}^{0} \right)
    \end{split}
\end{equation}
\vspace{-10pt}
\begin{gather}
\mathbf{\mathcal{X}}^{0} = 
\frac{-\mathbf{z}^{0} - 
h \left(\mathbf{\mathbf{x}} - 
\hat{\mathbf{\mathbf{x}}}\right)}{\sigma} , 
\textrm{ } \mathbf{\mathcal{X}}^{1} 
= \frac{- \mathbf{z}^{1} - 
h \left(\mathbf{\mathbf{x}} - 
\hat{\mathbf{\mathbf{x}}}\right)}{\sigma}
\end{gather}
% \end{equation}
where $Q(\mathbf{\mathbf{x}})$ is the Gaussian tail probability and $\sigma$ is 
the scalar of noise standard deviation that normalizes the measurement noise. 
The MAP problem(\autoref{MAP_problem}) is then equivalent to minimizing the following cost function:
\begin{equation}
    \begin{split}
    \mathcal{L\left(\mathbf{x}\right)} &= \left\|  \mathbf{x}_{k} \boxminus 
     \hat{\mathbf{x}}_{k} \right\|_{\mathbf{P}_{k}}^{2} 
    + \sum_{i=1}^{m}  2ln\left(Q \left( \mathbf{\mathcal{X}}^{1}_{i} \right) - 
    Q \left( \mathbf{\mathcal{X}}^{0}_{i} \right)\right) 
    \\
    &= 
     \left\|  \mathbf{x}_{k} \boxminus 
     \hat{\mathbf{x}}_{k} \right\|_{\mathbf{P}_{k}}^{2} 
    + \sum_{i=1}^{m} \left\| \mathbf{H}_{i}^{'} \tilde{\mathbf{x}} + z_{i}^{\prime} \right\|_{\mathbf{R}^{'}_{k}}^{2}
    \end{split}\label{eq_QMAP}
\end{equation}

By performing Taylor expansion on \autoref{eq_QMAP} and the assumption of quantized
measurement-based posterior PDFs will still be close to bell-shaped Gaussian PDFs\cite{msechu2008decentralized,trawny2009cooperative}
 the above equation can be solved using Kalman filtering\cite{xu2022fast,peng2024quantized}:
\begin{equation}
    \begin{aligned}
\mathbf{K} &=  \left({\mathbf{H^{'}}}^T \mathbf{R^{'}} \mathbf{H^{'}} + \mathbf{P}^{- 1} \right)
\mathbf{{H^{'}}^{T}}\mathbf {{R}^{'}}^{-1}
,
    \\
    \overline{\mathbf{x}}_{k}^{} &= \hat{\mathbf{x}}_{k}^{} \boxplus 
    \left(- \mathbf{K} \mathbf{z}^{'}_{k} - \left(\mathbf{I} - 
    \mathbf{K} \mathbf{H^{'}}\right) \left(\mathbf{J^{'}}
    ^{- 1}\right) \tilde{\mathbf{x}}_{k}\right)
    % \left(\hat{\mathbf{x}}_{k}^{} \boxminus 
        % \tilde{\mathbf{x}}_{k}\right)\right)
    \end{aligned}
\end{equation} % ?迭代记得改
Unlike the iterative approach \cite{trawny2009cooperative}, our method is executed in a single pass, eliminating the need for multiple iterations. For further technical details, we refer the reader to the relevant works \cite{msechu2008decentralized, peng2024quantized}.

% \begin{figure}[t]
%     \centering
%     \includegraphics[width=1\linewidth]{figure/directly quantization.png}
%     \caption{Conventional INT-8 Compression directly quantization the raw point coordinates(left), disrupting planar feature associations and causing failure due to accumulated geometric errors while QLIO (Right) only quantify the residual measurements}
 
%     \label{fig_rqvector}     
% \end{figure}
\begin{figure*}[t]
    \centering
    \includegraphics[width=1\linewidth]{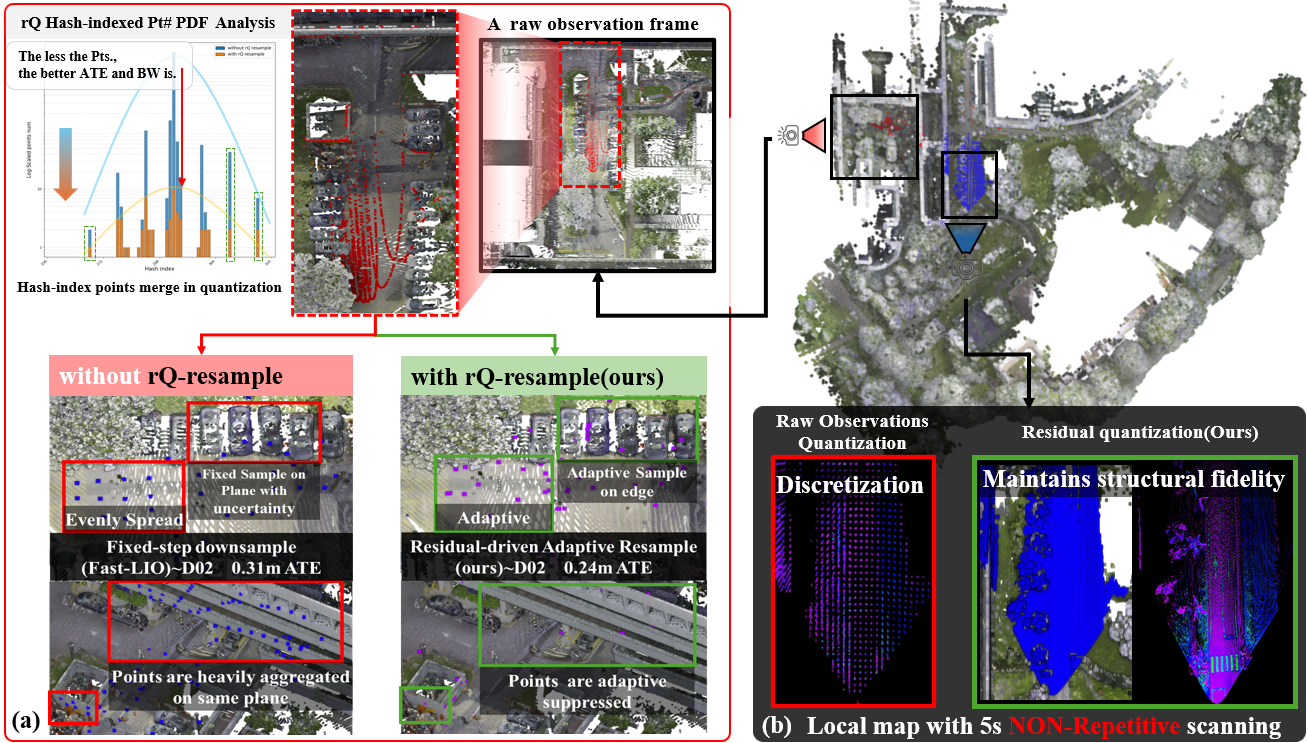}
        \vspace{-19pt}
    \caption{The test result on MCD-NTU dataset. (a) Conventional methods exhibit uniform planar point distributions, while our rQ-resample strategy preserves residual coherence, achieving lower ATE with fewer points. (b) Standard INT8 compression to the raw observation (left) induces discrete local map errors and causes point-plane association failure, whereas residual-aware quantization maintains structural fidelity.}
    \label{fig_567} 
    \vspace{-20pt}
\end{figure*}

\section{Experimental Results}
\subsection{Quantization Precision Analysis}
% Considering the correspondence between the main processor and the co-processor's 
% codebooks, as well as the encoding and decoding efficiency, 
% we adopt a simple and efficient fixed-point quantization sampling method. 
We adopt a simple and efficient fixed-point quantization
for the LiDAR point $\mathbf{p}$, the residual vector $\mathbf{n}$ 
and the meansurment ${\mathbf{z}}$, which maintains a static codebook configuration. 
We use $(-{r}_{max},{r}_{max})$ to represent the maximum quantization range of the LiDAR point (e.g. ${r}_{max}$=200m) and filter out 
the residual norm greater than the threshold ${r}_{thr}$ (e.g. 0.04 cm), meaning the residual space is truncated into a spherical space 
(see figure \ref{fig_rqvector}).
Let ${l}_{p}$, ${l}_{n}$ and ${l}_{z}$ as the quantization number of bits, fixed-point multi-bit $\mathbf{p}_{}^{\mathbf{\mathit{q}}}$, $\mathbf{n}_{}^{\mathbf{\mathit{q}}}$, ${\mathbf{z}^{q}}$ and the required bits number with out rQ $N$ can represent as:
\begin{equation}
    \begin{aligned}
    \mathbf{p}^{\mathbf{\mathit{q}}} &= 
    \mathbf{f}_{{p}}^{\mathbf{\mathit{q}}} 
    \left(\mathbf{p}\right) = 
    (p_{1}^{q} , p_{2}^{q} , p_{3}^{q}), \
    \mathbf{n}_{}^{\mathbf{\mathit{q}}} = 
    \mathbf{f}_{{n}}^{\mathbf{\mathit{q}}} 
    \left(\mathbf{n}\right) = 
    (n_{1}^{q} , n_{2}^{q} , n_{3}^{q})
    \\
     p_j^q &= \left\lfloor \dfrac{p_j + r_{max}}{2^{(1 - l_p)}r_{max}} \right\rfloor \cdot 2^{(1 - l_p)}r_{max} - r_{max} + 2^{( -l_p)}r_{max}
    \\
    n_j^q &= \left\lfloor \dfrac{n_j + r_{thr}}{2^{(1 - l_n)}r_{thr}} \right\rfloor \cdot 2^{(1 - l_n)}r_{thr} - r_{thr} + 2^{(-l_n)}r_{thr}
    \\
    \mathbf{z}^{q} &= \left\lfloor \dfrac{\mathbf{z}}
    {2^{(- l_{\mathbf{\mathit{z}}} )}{r}_{thr}}
     \right\rfloor \cdot 2^{(-l_{\mathbf{\mathit{z}}})} {r}_{thr}
     + 2^{(- l_{\mathbf{\mathit{z}}} - 1)}{r}_{thr}
     \\ 
     N &= 3l_{\mathbf{\mathit{p}}} + 3l_{\mathbf{\mathit{n}}} +l_{\mathbf{\mathit{z}}}, j=1,2,3.
    \end{aligned}
\end{equation}
where \scalebox{0.8 }{$\left\lfloor \dfrac{p_j + r_{max}}{2^{(1 - l_p)}r_{max}} \right\rfloor$},
  \scalebox{0.8}{$ \left\lfloor \dfrac{n_j + r_{thr}}{2^{(1 - l_n)}r_{thr}} \right\rfloor$, 
   \scalebox{0.9}{$\left\lfloor \dfrac{\mathbf{z}}
    {2^{(- l_{\mathbf{\mathit{z}}} )}{r}_{thr}}
     \right\rfloor$ }} 
can then compressed to the binary bit stream based on rQ-vector grouping
compressed strategy 
\autoref{compressed_stragy}.
    % \begin{figure}
    %     % width=0.5\textwidth
    %     % [width=10cm]
    %     \includegraphics[width=0.5\textwidth]{figure/distribution.pdf}              % [width=32cm]
    %     \caption{Comparison of Point Distribution With and Without Resampling.}
    %     \vspace{-20pt} % 调整段落和图片之间的间距
    %     \label{fig_result}     
    % \end{figure}
    % \begin{figure}
    %     % width=0.5\textwidth
    %     % [width=10cm]
    %     \includegraphics[width=1\linewidth]{figure/figure_resample.png}           
    %     \caption{The point observation comparison between voxel-based downsampling (left) and 
    %     rQ-based resampling (right). The voxel stride size was calibrated to achieve equivalent point density with the rQrs method.}
    %     \vspace{-20pt} % 调整段落和图片之间的间距
    %     \label{fig_resample}     
    % \end{figure}

\subsection{Real World Experiment}
We evaluate the proposed QLIO on the MCD-NTU dataset\cite{nguyen2024mcd}. 
The MCD-NTU dataset was collected using an ATV within the NTU campus
with a solid-state hybrid LiDAR (Livox-Mid70, Mid), 
a mechanical LiDAR (Ouster128, OS) and an external IMU sensor(VN100), including 6 sequnces. 
Our system is built based on FastLIO\cite{xu2022fast}.
We first conduct a comparative evaluation between FastLIO and our residual-quantized (rQ) vector-based resampling strategy (rQrs) employing fixed hash hierarchy levels ($l_n=3$, equivalent to 512 hash buckets). The ``X'' means failed below and the average absolute trajectory error(ATE) results are shown in the \autoref{table_ATE_g1}. It can be observed that the accuracy of FastLIO has been significantly improved after the application of rQrs, which is attributed to the resampling of the point observation and the residual, as shown in the \autoref{fig_567}. 
\begin{table}
    \centering
    \caption{ATE in meters for different methods and sequences.}\label{table_ATE_g1}
    \begin{tabular}{c c c c c c c c}
        \toprule
        Method/Seq.         & \textbf{D01} & \textbf{D02} & \textbf{D10} & \textbf{N04} & \textbf{N08} & \textbf{N13} \\%& \textbf{bits/scans} & measuretime \\ 
        \midrule
    \textbf{OS}           & 1.48           & 0.27           & 2.08           & 1.59            & 2.05           & \underline{1.29}  \\  %&     &  \\ %\hline
    \textbf{OS-rQrs}      & 1.28           & \textbf{0.23}  & \textbf{1.28}  & 1.25            & 1.62            & \textbf{0.89} \\ % &      &  \\ %\hline
\textbf{Mid}        & \underline{1.09}      & 0.31          & 1.63           & \underline{0.85} & \underline{1.54} & 1.88   \\ % &    &  \\ %\hline
    \textbf{Mid-rQrs}   & \textbf{0.91}   & \underline{0.24}& \underline{1.47}& \textbf{0.49}   & \textbf{1.33}  &  2.33    & \\  % &\\ %\hline
    \textbf{OS-dacro(8)}  & X               & X             & X               & X               & X               & X \\  %  &      &    \\ %\hline
    \textbf{Mid-INT(8)} & X               & 15.75           & X               & X               & X               & X   \\  %  &      &  \\ 
    \bottomrule
    \end{tabular}
\end{table}

\begin{table}
    \centering
    \caption{ for different methods and sequences.}\label{table_costtime1}
    \vspace{-10pt}
    \begin{tabular}{c c c c c c}
        \toprule
        Method/Seq.          & d-c cost/ms & p.r   & bits/mea. & avg.num & \\%& \textbf{bits/scans} & measuretime \\ 
        \midrule
\textbf{OS}           &  0              &  1          &     96       &     15020.65   &  \\  %&     &  \\ %\hline
\textbf{OS-rQrs}      & 0.59             &  1          &     96       &      \underline{566.07}   & \\ % &      &  \\ %\hline
    \textbf{Mid}        &    0          &  1          &     96       &   2799.16   &     \\ % &    &  \\ %\hline
    \textbf{Mid-rQrs}   & \underline{0.23}           &    1         &     96       &     \textbf{248.98}  &     \\  % &\\ %\hline
\textbf{OS-dacro(8)}    &  1868.25       & \textbf{3.98}       &      \underline{24.12}       &    16555.97  &    \\  %  &      &    \\ %\hline
    \textbf{Mid-INT(8)} & \textbf{0.14}           & \underline{4}           &     \textbf{24}        &     2665.32 &   \\  %  &      &  \\ 
    \bottomrule
    \end{tabular}
    \vspace{-15pt}
\end{table}
The fixed-step downsampling method arranges observations more regularly(see \autoref{fig_567}.a ``without rQ-resample''), yet induces directional residual concentration bias through its fixed sampling interval, resulting in observation direction imbalance (as per distribution analysis). However, our method resolves this imbalance, achieving measurement uniformity with fewer sampling points while maintaining optimization stability.
% The fixed step based downsampling method let the observation more regularly arranged and it discards those with different residual directions (as shown in \autoref{fig_567}.a ``without rQ-resample''). At the same time, due to the use of a fixed step size, more observations tend to concentrate in similar observation directions, resulting in an observation directions imbalance (as shown in the distribution analysis). However, rQrs fundamentally resolves this distribution imbalance, achieving enhanced measurement uniformity through fewer sampling points in state estimation while maintaining optimization stability.
 % The average absolute trajectory error(ATE) results are shown in the \autoref{table_ATE_g1}. 
 
 In addition, we implemented two conventional point cloud compression methods within our LiDAR-Inertial Odometry (LIO) pipeline to simulate centralized processing with fully compressed observations. To our knowledge, conventional compression methods are rarely tested in such tightly coupled LIO frameworks (most existing evaluations focus on pure LiDAR odometry like LOAM). Specifically, we applied INT8 Quantization and 8-bit Dacro two distinct compression techniques. Both approaches maintained identical post-processing pipelines with the original LIO algorithm where p.r (point cloud compression ratio), d-c (decompress-compress) cost time, and avg.num (average number of points per scan included in state estimation) were measured under varying compression configurations. Surprisingly, both methods exhibited significant performance degradation:
INT8 fails because of the precision loss from point value truncation destroyed planar continuity, as shown in \autoref{fig_567}.b, where original smooth surfaces became discrete grids. This fundamentally disrupted point-to-plane data association. Dacro limits because of the excessive compression/decompression latency (see \autoref{table_costtime1}) caused cumulative IMU integration drift during processing gaps, ultimately compromising state estimation. It should be noted that the observed INT8 quantization-induced plane matching failures may stem from implementation specifics rather than inherent limitations of INT8 quantization. Alternative quantization strategies (e.g., log-scale quantization, variable-bit quantization) could theoretically resolve this, though our exhaustive experiments with these approaches revealed suboptimal performance-all variants required extensive hyperparameter tuning with additional parameters (detailed ablation studies will be documented in the supplementary material) thereby introducing prohibitive complexity in benchmark control. Consequently, we adopted the deterministic min-max INT8 quantization scheme to ensure reproducibility, sacrificing marginal accuracy for experimental stability.
Notably, these approaches represent common paradigms for edge-device state estimation and we will then introduce our targeted solution QLIO framework.
\begin{table}[H]
    \centering
    \renewcommand{\arraystretch}{1.3}
    \caption{ATE meters/rads in different ${l}_{n}$ influences}\label{table_gen2}
    \vspace{-10pt}
    \setlength{\tabcolsep}{3pt}
    \begin{tabular}{c p{0.2cm} c c c c c c}
    \hline
    \hline
        \toprule
        M/O  &${l}_{n}$ & \textbf{D01} & \textbf{D02} & \textbf{D10} & \textbf{N04} & \textbf{N08} & \textbf{N13} \\ 
        \midrule
        \multirow{3}{*}{OS}  
        & 1  & X                            & 1.21/\underline{0.02}             & X                 & X             & X               & X            \\
        & 2  & 1.57/\textbf{0.02}           & \textbf{0.25}/\textbf{0.01}       & \underline{1.39}/\underline{0.03}         & 1.37/\textbf{0.03}     & \underline{1.54}/\textbf{0.02} & \textbf{0.86}/\textbf{0.04}   \\
        & 3  & \underline{1.41}/\textbf{0.02}  & \underline{0.29}/\textbf{0.01} & \textbf{1.36}/\underline{0.03} & 1.36/\textbf{0.03} & 1.82/\textbf{0.02} & \underline{0.91}/\textbf{0.04}  \\\hline
        \multirow{3}{*}{Mid}  
        & 1 & X                             & X                                 & X                 & X           & X             & X    \\  
        & 2  & 1.96/\underline{0.03}        & 0.56/\underline{0.02} & 1.57/\textbf{0.02} & \underline{1.32}/\textbf{0.03}    & 3.89/\textbf{0.02} & X      \\
        & 3 & \textbf{1.23}/\underline{0.03}  & \underline{0.29}/\underline{0.02}           & 1.41/\textbf{0.02} & \textbf{0.62}/\textbf{0.03} & \textbf{1.40}/\textbf{0.02} & X       \\
        \bottomrule
    \end{tabular}
    \vspace{-10pt}
\end{table}
\begin{table}[H]
    \centering
    \renewcommand{\arraystretch}{1.3}
    \caption{ATE in meters/rads for different methods and sequences using differnt ${l}_{z}$.}\label{table_gen3}
     \vspace{-10pt}
    \setlength{\tabcolsep}{3pt} % 调整列间距，默认是6pt
    \begin{tabular}
        {c p{0.2cm} c c c c c c}
        \hline
        \hline
        \toprule
            M/O                                    &${l}_{z}$& \textbf{D01}  & \textbf{D02} & \textbf{D10} & \textbf{N04}  & \textbf{N08} & \textbf{N13}        \\
        \midrule
    \multirow{4}{*}{\hspace{0pt}OS}        
& 1 & 1.37/\textbf{0.02}             &    \textbf{0.24}/\textbf{0.01}     & \textbf{1.32}/\underline{0.03}        & 1.40/0.03                      & 1.76/\textbf{0.02}         & \underline{0.88}/\textbf{0.04}  \\
& 2 & 1.57/\textbf{0.02}             & \textbf{0.24}/\textbf{0.01}     & \underline{1.36}/\underline{0.03}         & 1.32/\textbf{0.03}         & \underline{1.58}/\textbf{0.02}         & \textbf{0.86}/\textbf{0.04}  \\
& 3 & 1.30/\textbf{0.02}             & \underline{0.25}/\textbf{0.01}  & 1.44/\underline{0.03}                     & 1.41/\textbf{0.03}         & 1.90/\textbf{0.02}         & \underline{0.88}/\textbf{0.04}  \\
& 4 & \textbf{1.23}/\textbf{0.02}    & \textbf{0.24}/\textbf{0.01}     & 1.51/\underline{0.03}                        & 1.43/\textbf{0.03}         & 1.79/\textbf{0.02}         & \textbf{0.86}/\textbf{0.04}  \\\hline
    \multirow{4}{*}{\hspace{0pt}Mid}               
& 1 & 1.34/\textbf{0.02}             & 0.43/\underline{0.02}         & 1.93/\textbf{0.02}                      & 1.75/0.04                             & 3.09/\textbf{0.02}         &     X       \\
& 2 & 1.99/\textbf{0.02}             & 0.29/\underline{0.02}        & 1.97/\textbf{0.02}                         & \underline{1.23}/\textbf{0.03}            & 3.28/\textbf{0.02}         &     X       \\
& 3 & 1.36/\textbf{0.02}             & 0.32/\underline{0.02}         & 1.38/\textbf{0.02}                         & \textbf{1.21}/0.04                & \textbf{\textbf{1.56}}/\textbf{0.02} &     X       \\
& 4 & \underline{1.26}/\textbf{0.02}             & 0.31/\underline{0.02}         & 1.57/\textbf{0.02}                         & 1.26/0.04             & 2.46/\textbf{0.02}         &     X      \\
    \bottomrule
        \end{tabular}
         \vspace{-10pt}
    \end{table}

 We first performed comparative ablation studies on residual component quantization bits ${l}_{n}$ versus residual-vector quantization bits ${l}_{z}$ (which may also directly determine the number of the hash bucket)
 systematically varying each parameter from 1-bit to 3-bit configurations under rQrs. This parametric sweep enabled empirical analysis of accuracy variations under progressive quantization refinement.
 Our experiments demonstrate that residual vector quantization with ${l}_{n}=1$ consistently induces system failure across both OS and MID platforms, indicating catastrophic sensitivity to insufficient quantization precision. The experimental results of increasing the quantization level reveal distinct platform-dependent characteristics: ${l}_{n}$ demonstrates negligible accuracy variation on OS, whereas MID exhibits consistent precision improvement that correlates with its unique multi-echo radar architecture. As the residual-vector quantization bits ${l}_{z}$ increases, both MID and OS demonstrate minimal variation across all test sequences, with the notable exception of the N13 sequence (as shown in \autoref{table_gen3}).

On the N13 sequence involving rapid 180° rotations, we observed MID70 divergence. This is attributed to mid70’s single-frame observation mechanism, which differs from Ouster in that it fails to capture a complete spherical point cloud. The incomplete spatial sampling leads to concentrated distributions of residual vectors after low-bit quantization, significantly disrupting original data associations. As we progressively increased quantization bits, system accuracy recovered accordingly. Precision returned to baseline levels when ${l}_{n}=3$. This demonstrates that QLIO requires higher bit allocation under degraded conditions to maintain system stability. However, even when increasing to 4-bit quantization (compared to the original 32-bit float representation), this tradeoff remains worthwhile for devices with stringent communication bandwidth requirements.
\begin{table}[H]
\centering
    \caption{ATE result in different combination of ${l}_{z}$ and ${l}_{n}$ using Mid70 in N13.}\label{table_gen3}
    \setlength{\tabcolsep}{3pt} % 调整列间距，默认是6pt
    \begin{tabular}
        {c c c c}
        \toprule
\textbf{N13-MID} & ${l}_{n}=1$ & ${l}_{n}=2$ & ${l}_{n}=3$\\
        \midrule
${l}_{z}=1$         & X & X          & X              \\
${l}_{z}=2$         & X & X          & 1.31/\textbf{0.04}  \\
${l}_{z}=3$         & X & 15.54/0.14 & \textbf{1.25}/\textbf{0.04}  \\
${l}_{z}=4$         & X & 4.14/0.06  & 1.45/\textbf{0.04} \\
\bottomrule
\end{tabular}
\vspace{-10pt}
\end{table}

\begin{table*}
    \centering
    \renewcommand{\arraystretch}{1.1}
    \caption{ATE in meters/rads and the bits/Bytes occupation‌ per measurements for ${l}_{p}$ and sequences.}\label{table_gen4}
    \vspace{-10pt}
    \begin{tabular}{c c c c c c c c c c c c c c c}
        \toprule
    \textbf{seq}            & \textbf{rQrs} & ${l}_{p}=3$ & ${l}_{p}=4$ & ${l}_{p}=5$ & ${l}_{p}=6$ & ${l}_{p}=7$ & ${l}_{p}=8$ & ${l}_{p}=9$ & ${l}_{p}=10$ & ${l}_{p}=11$ & ${l}_{p}=12$ \\
    \midrule
     
    \multirow{2}{*}{D01}    & \textbf{w/o} & 2.95/0.04 & 2.10/0.04 & 1.76/0.04 & 1.62/0.04 & 1.59/0.04 & 1.63/0.04 & 1.61/0.04 & 1.57/0.04 & 1.64/0.04 & 1.68/0.04 \\
                            & \textbf{w/} & 1.98/0.03 & 1.74/0.03 & 1.51/0.03 & 1.52/0.03 & 1.40/0.03 & 1.48/0.03 & 1.46/0.03 & 1.55/0.03 & 1.44/0.03 & 1.51/0.03 \\\hline
    \multirow{2}{*}{D02}    & \textbf{w/o} & 0.33/0.01 & 0.31/0.01 & 0.32/0.01 & 0.31/0.01 & 0.31/0.01 & 0.31/0.01 & 0.31/0.01 & 0.32/0.01 & 0.31/0.01 & 0.31/0.01 \\
                            & \textbf{w/} & 0.24/0.01 & 0.24/0.01 & 0.25/0.01 & 0.26/0.01 & 0.25/0.01 & 0.25/0.01 & 0.25/0.01 & 0.25/0.01 & 0.26/0.01 & 0.26/0.01 \\\hline
    \multirow{2}{*}{D10}    & \textbf{w/o} & 2.74/0.04 & 2.52/0.04 & 2.74/0.04 & 2.76/0.04 & 2.63/0.04 & 2.71/0.04 & 2.69/0.04 & 2.66/0.04 & 2.61/0.04 & 2.58/0.04 \\
                            & \textbf{w/} & 1.18/0.04 & 1.06/0.04 & 1.31/0.04 & 1.27/0.04 & 1.36/0.04 & 1.34/0.04 & 1.30/0.04 & 1.30/0.04 & 1.42/0.04 & 1.30/0.04 \\\hline
    \multirow{2}{*}{N04}    & \textbf{w/o} & 1.60/0.03 & 1.61/0.03 & 1.64/0.03 & 1.64/0.03 & 1.67/0.03 & 1.66/0.03 & 1.70/0.03 & 1.67/0.03 & 1.67/0.03 & 1.65/0.03 \\
                            & \textbf{w/} & 1.21/0.03 & 1.27/0.03 & 1.31/0.03 & 1.41/0.03 & 1.31/0.03 & 1.37/0.03 & 1.41/0.03 & 1.39/0.03 & 1.39/0.03 & 1.40/0.03 \\\hline
    \multirow{2}{*}{N08}    & \textbf{w/o} & 2.62/0.02 & 2.76/0.02 & 3.17/0.02 & 2.91/0.02 & 3.08/0.02 & 2.98/0.02 & 2.98/0.02 & 3.07/0.02 & 3.06/0.02 & 2.89/0.02 \\
                            & \textbf{w/} & 2.07/0.02 & 1.53/0.02 & 1.67/0.02 & 1.85/0.02 & 1.68/0.02 & 1.48/0.02 & 1.54/0.02 & 1.84/0.02 & 1.56/0.02 & 1.75/0.02 \\\hline
    \multirow{2}{*}{N13}     
    &\textbf{w/o} & 1.41/0.05 & 1.34/0.05 & 1.30/0.05 & 1.30/0.05 & 1.29/0.05 & 1.29/0.05 & 1.29/0.05 & 1.30/0.05 & 1.29/0.05 & 1.29/0.05\\  & \textbf{w/} & 0.89/0.05 & 0.85/0.05 & 0.86/0.05 & 0.87/0.05 & 0.88/0.05 & 0.88/0.05 & 0.89/0.05 & 0.87/0.05 & 0.87/0.05 & 0.87/0.05\\    
                            \hline
    \multirow{2}{*}{avg.b/B} &\textbf{w/o} & 20/2.50 & 23/2.87 & 26/3.25 & 29/3.62 & 32/4.00 & 35/4.37 & 38/4.75 & 41/5.12 & 44/5.50 & 47/5.87 \\ 
    & \textbf{w/} & 5.84/0.73 &	6.84/0.85 & 7.84/0.98	& 8.84/1.10& 	9.84/1.23&	10.84/1.35&	11.84/1.48	&12.84/1.60&	13.84/1.73	&14.84/1.85 \\
    \bottomrule
    \end{tabular}
    \vspace{-10pt}
\end{table*}

Based on our prior experiments, we fixed the 3-bit residual vector quantization and 2-bit residual quantization schemes(${l}_{n}=3$, ${l}_{z}=2$) then conducted systematic tests on LiDAR point quantization bits ranging from 3 to 12 bits using Ouster. The corresponding ATE meter/rads results, per-observation bit counts, and byte consumption are detailed in\autoref{table_gen4} .The results demonstrate that implementing the rQrs strategy significantly enhances precision while drastically reducing observation counts. This efficiency gain stems from our group compression protocol - points sharing same rQ-vector are packaged together during compression, thereby reducing the cost required for rQ-vector processing.
\begin{figure}[t]
        % width=0.5\textwidth
        % [width=10cm]
        \includegraphics[width=0.5\textwidth]{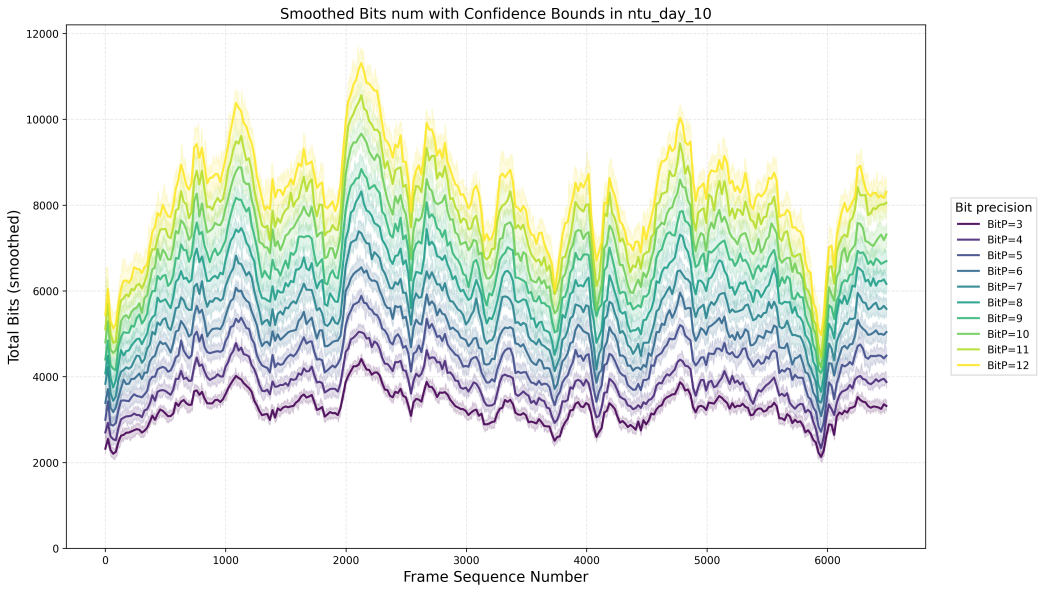}              % [width=32cm]
        \vspace{-21pt}
        \caption{The total bits per frame in D10 seq.}
        \vspace{-15pt} % 调整段落和图片之间的间距
        \label{fig_D10result}     
    \end{figure}   
Surprisingly, significant fluctuations in the quantization bit number did not lead to substantial accuracy loss. Notably, progressive reduction of quantization bits from 12 to 3 bits reveals a precision-bandwidth trade-off: 3-bit quantization achieves minimal measurement overhead (5.84bits per mea.)but also the minimal precision while more bit quantization achieves more precision gain. 
Crucially, even with 12-bit quantization under rQrs(14.84bits per mea.), total per-frame bandwidth consumption on the MCD-D10 dataset remains below 1.5kB(as shown in \autoref{fig_D10result}). This translates to $<$1.2Mbps total bandwidth usage, enabling practical transmission via edge-device-friendly protocols like UART/SPI (typically supporting 1-10Mbps rates).

\section{Conclusion and Future Work}

QLIO redefines LIO deployment on edge devices by integrating multi-processor quantization, adaptive resampling, and extreme bandwidth reduction. It enables real-time LiDAR state estimation on low-power platforms such as drones and humanoid robots. Unlike 2D visual feature quantization methods like QVIO, QLIO introduces structured 3D point-to-plane quantization and a LiDAR-specific MAP-based estimator. The experimental results show a 15.1× per-observation bandwidth reduction while preserving localization accuracy, demonstrating the need for sensor-specific quantization strategies in LiDAR odometry.

However, QLIO currently relies on heuristic quantization parameters, which may \textbf{limit} adaptability to dynamically changing environments. Additionally, its \textbf{multi-processor architecture}, while improving efficiency, introduces hardware dependencies that require careful integration with different computing platforms. 
Our future research will prioritize the development of an \textbf{intelligent parameter optimization framework}, such as implementing a reinforcement learning-based architecture with multi-head attention mechanisms for adaptive point cloud feature selection. This approach will systematically replace heuristic parameter design in QLIO with bandwidth-aware optimization, particularly improving spectral efficiency in SLAM applications through adaptive data compression strategies based on communication condition monitoring.

\balance
\bibliographystyle{IEEEtran}
\bibliography{IEEEfull}

\begin{thebibliography}{10}
\providecommand{\url}[1]{#1}
\csname url@samestyle\endcsname
\providecommand{\newblock}{\relax}
\providecommand{\bibinfo}[2]{#2}
\providecommand{\BIBentrySTDinterwordspacing}{\spaceskip=0pt\relax}
\providecommand{\BIBentryALTinterwordstretchfactor}{4}
\providecommand{\BIBentryALTinterwordspacing}{\spaceskip=\fontdimen2\font plus
\BIBentryALTinterwordstretchfactor\fontdimen3\font minus \fontdimen4\font\relax}
\providecommand{\BIBforeignlanguage}[2]{{%
\expandafter\ifx\csname l@#1\endcsname\relax
\typeout{** WARNING: IEEEtran.bst: No hyphenation pattern has been}%
\typeout{** loaded for the language `#1'. Using the pattern for}%
\typeout{** the default language instead.}%
\else
\language=\csname l@#1\endcsname
\fi
#2}}
\providecommand{\BIBdecl}{\relax}
\BIBdecl

\bibitem{yuan2024mmaud}
S.~Yuan, Y.~Yang, T.~H. Nguyen, T.-M. Nguyen, J.~Yang, F.~Liu, J.~Li, H.~Wang, and L.~Xie, ``Mmaud: A comprehensive multi-modal anti-uav dataset for modern miniature drone threats,'' in \emph{2024 IEEE International Conference on Robotics and Automation (ICRA)}, 2024, pp. 2745--2751.

\bibitem{he2024omnih2o}
T.~He, Z.~Luo, X.~He, W.~Xiao, C.~Zhang, W.~Zhang, K.~M. Kitani, C.~Liu, and G.~Shi, ``Omnih2o: Universal and dexterous human-to-humanoid whole-body teleoperation and learning,'' \emph{Conference on Robot Learning (CoRL)}, 2024.

\bibitem{li2024hcto}
J.~Li, S.~Yuan, M.~Cao, T.-M. Nguyen, K.~Cao, and L.~Xie, ``Hcto: Optimality-aware lidar inertial odometry with hybrid continuous time optimization for compact wearable mapping system,'' \emph{ISPRS Journal of Photogrammetry and Remote Sensing}, vol. 211, pp. 228--243, 2024.

\bibitem{li2025helmetposer}
J.~Li, Q.~Leng, J.~Liu, X.~Xu, T.~Jin, M.~Cao, T.-M. Nguyen, S.~Yuan, K.~Cao, and L.~Xie, ``Helmetposer: A helmet-mounted imu dataset for data-driven estimation of human head motion in diverse conditions,'' in \emph{Proceedings of the IEEE International Conference on Robotics and Automation (ICRA)}, 2025.

\bibitem{xu2022fast}
W.~Xu, Y.~Cai, D.~He, J.~Lin, and F.~Zhang, ``Fast-lio2: Fast direct lidar-inertial odometry,'' \emph{IEEE Transactions on Robotics}, vol.~38, no.~4, pp. 2053--2073, 2022.

\bibitem{bai2022faster}
C.~Bai, T.~Xiao, Y.~Chen, H.~Wang, F.~Zhang, and X.~Gao, ``Faster-lio: Lightweight tightly coupled lidar-inertial odometry using parallel sparse incremental voxels,'' \emph{IEEE Robotics and Automation Letters}, vol.~7, no.~2, pp. 4861--4868, 2022.

\bibitem{nguyen2024eigen}
T.-M. Nguyen, X.~Xu, T.~Jin, Y.~Yang, J.~Li, S.~Yuan, and L.~Xie, ``Eigen is all you need: Efficient lidar-inertial continuous-time odometry with internal association,'' \emph{IEEE Robotics and Automation Letters}, vol.~9, no.~6, pp. 5330--5337, 2024.

\bibitem{shan2020lio}
T.~Shan, B.~Englot, D.~Meyers, W.~Wang, C.~Ratti, and D.~Rus, ``Lio-sam: Tightly-coupled lidar inertial odometry via smoothing and mapping,'' in \emph{2020 IEEE/RSJ international conference on intelligent robots and systems (IROS)}.\hskip 1em plus 0.5em minus 0.4em\relax IEEE, 2020, pp. 5135--5142.

\bibitem{nguyen2021miliom}
T.-M. Nguyen, S.~Yuan, M.~Cao, L.~Yang, T.~H. Nguyen, and L.~Xie, ``Miliom: Tightly coupled multi-input lidar-inertia odometry and mapping,'' \emph{IEEE Robotics and Automation Letters}, vol.~6, no.~3, pp. 5573--5580, 2021.

\bibitem{shan2021lvi}
T.~Shan, B.~Englot, C.~Ratti, and D.~Rus, ``Lvi-sam: Tightly-coupled lidar-visual-inertial odometry via smoothing and mapping,'' in \emph{2021 IEEE international conference on robotics and automation (ICRA)}.\hskip 1em plus 0.5em minus 0.4em\relax IEEE, 2021, pp. 5692--5698.

\bibitem{lim2023adalio}
H.~Lim, D.~Kim, B.~Kim, and H.~Myung, ``Adalio: Robust adaptive lidar-inertial odometry in degenerate indoor environments,'' in \emph{2023 20th International Conference on Ubiquitous Robots (UR)}.\hskip 1em plus 0.5em minus 0.4em\relax IEEE, 2023, pp. 48--53.

\bibitem{nguyen2023slict}
T.-M. Nguyen, D.~Duberg, P.~Jensfelt, S.~Yuan, and L.~Xie, ``Slict: Multi-input multi-scale surfel-based lidar-inertial continuous-time odometry and mapping,'' \emph{IEEE Robotics and Automation Letters}, vol.~8, no.~4, pp. 2102--2109, 2023.

\bibitem{vizzo2023kiss}
I.~Vizzo, T.~Guadagnino, B.~Mersch, L.~Wiesmann, J.~Behley, and C.~Stachniss, ``Kiss-icp: In defense of point-to-point icp--simple, accurate, and robust registration if done the right way,'' \emph{IEEE Robotics and Automation Letters}, vol.~8, no.~2, pp. 1029--1036, 2023.

\bibitem{chen2023direct}
K.~Chen, R.~Nemiroff, and B.~T. Lopez, ``Direct lidar-inertial odometry: Lightweight lio with continuous-time motion correction,'' in \emph{2023 IEEE international conference on robotics and automation (ICRA)}.\hskip 1em plus 0.5em minus 0.4em\relax IEEE, 2023, pp. 3983--3989.

\bibitem{chen2024ig}
Z.~Chen, Y.~Xu, S.~Yuan, and L.~Xie, ``ig-lio: An incremental gicp-based tightly-coupled lidar-inertial odometry,'' \emph{IEEE Robotics and Automation Letters}, vol.~9, no.~2, pp. 1883--1890, 2024.

\bibitem{ji2024lio}
X.~Ji, S.~Yuan, P.~Yin, and L.~Xie, ``Lio-gvm: an accurate, tightly-coupled lidar-inertial odometry with gaussian voxel map,'' \emph{IEEE Robotics and Automation Letters}, vol.~9, no.~3, pp. 2200--2207, 2024.

\bibitem{wu2024lio}
Y.~Wu, T.~Guadagnino, L.~Wiesmann, L.~Klingbeil, C.~Stachniss, and H.~Kuhlmann, ``Lio-ekf: High frequency lidar-inertial odometry using extended kalman filters,'' in \emph{2024 IEEE International Conference on Robotics and Automation (ICRA)}.\hskip 1em plus 0.5em minus 0.4em\relax IEEE, 2024, pp. 13\,741--13\,747.

\bibitem{eisoldt2022fully}
M.~Eisoldt, J.~Gaal, T.~Wiemann, M.~Flottmann, M.~Rothmann, M.~Tassemeier, and M.~Porrmann, ``A fully integrated system for hardware-accelerated tsdf slam with lidar sensors (hatsdf slam),'' \emph{Robotics and Autonomous Systems}, vol. 156, p. 104205, 2022.

\bibitem{lin2022r}
J.~Lin and F.~Zhang, ``R 3 live: A robust, real-time, rgb-colored, lidar-inertial-visual tightly-coupled state estimation and mapping package,'' in \emph{2022 International Conference on Robotics and Automation (ICRA)}.\hskip 1em plus 0.5em minus 0.4em\relax IEEE, 2022, pp. 10\,672--10\,678.

\bibitem{jin2024robust}
T.~Jin, X.~Xu, Y.~Yang, S.~Yuan, T.-M. Nguyen, J.~Li, and L.~Xie, ``Robust loop closure by textual cues in challenging environments,'' \emph{IEEE Robotics and Automation Letters}, vol.~10, no.~1, pp. 812--819, 2025.

\bibitem{Li2024PSS}
J.~Li, T.-M. Nguyen, S.~Yuan, and L.~Xie, ``Pss-ba: Lidar bundle adjustment with progressive spatial smoothing,'' in \emph{2024 IEEE/RSJ International Conference on Intelligent Robots and Systems (IROS)}, 2024, pp. 1124--1129.

\bibitem{Li2024graph}
J.~Li, T.-M. Nguyen, M.~Cao, S.~Yuan, T.-Y. Hung, and L.~Xie, ``Graph optimality-aware stochastic lidar bundle adjustment with progressive spatial smoothing,'' in \emph{arXiv preprint arXiv:2410.14565}, 2024.

\bibitem{ma2024mm}
Y.~Ma, J.~Xu, S.~Yuan, T.~Zhi, W.~Yu, J.~Zhou, and L.~Xie, ``Mm-lins: a multi-map lidar-inertial system for over-degenerate environments,'' \emph{IEEE Transactions on Intelligent Vehicles}, vol.~-, pp. 1--11, 2024.

\bibitem{yin2024outram}
P.~Yin, H.~Cao, T.-M. Nguyen, S.~Yuan, S.~Zhang, K.~Liu, and L.~Xie, ``Outram: One-shot global localization via triangulated scene graph and global outlier pruning,'' in \emph{2024 IEEE International Conference on Robotics and Automation (ICRA)}.\hskip 1em plus 0.5em minus 0.4em\relax IEEE, 2024, pp. 13\,717--13\,723.

\bibitem{Zhao2024adaptive}
C.~Zhao, K.~Hu, J.~Xu, L.~Zhao, B.~Han, K.~Wu, M.~Tian, and S.~Yuan, ``Adaptive-lio: Enhancing robustness and precision through environmental adaptation in lidar inertial odometry,'' \emph{IEEE Internet of Things Journal}, 2024.

\bibitem{rebecq2016evo}
H.~Rebecq, T.~Horstsch{\"a}fer, G.~Gallego, and D.~Scaramuzza, ``Evo: A geometric approach to event-based 6-dof parallel tracking and mapping in real time,'' \emph{IEEE Robotics and Automation Letters}, vol.~2, no.~2, pp. 593--600, 2016.

\bibitem{wang2020tartanair}
W.~Wang, D.~Zhu, X.~Wang, Y.~Hu, Y.~Qiu, C.~Wang, Y.~Hu, A.~Kapoor, and S.~Scherer, ``Tartanair: A dataset to push the limits of visual slam,'' in \emph{2020 IEEE/RSJ International Conference on Intelligent Robots and Systems (IROS)}.\hskip 1em plus 0.5em minus 0.4em\relax IEEE, 2020, pp. 4909--4916.

\bibitem{helmberger2022hilti}
M.~Helmberger, K.~Morin, B.~Berner, N.~Kumar, G.~Cioffi, and D.~Scaramuzza, ``The hilti slam challenge dataset,'' \emph{IEEE Robotics and Automation Letters}, vol.~7, no.~3, pp. 7518--7525, 2022.

\bibitem{nair2024hilti}
A.~D. Nair, J.~Kindle, P.~Levchev, and D.~Scaramuzza, ``Hilti slam challenge 2023: Benchmarking single+ multi-session slam across sensor constellations in construction,'' \emph{IEEE Robotics and Automation Letters}, 2024.

\bibitem{wang2017non}
C.~Wang, J.~Yuan, and L.~Xie, ``Non-iterative slam,'' in \emph{2017 18th International Conference on Advanced Robotics (ICAR)}.\hskip 1em plus 0.5em minus 0.4em\relax IEEE, 2017, pp. 83--90.

\bibitem{yang2024fast}
Z.~Yang, K.~Xu, S.~Yuan, and L.~Xie, ``A fast and light-weight noniterative visual odometry with rgb-d cameras,'' \emph{Unmanned Systems}, vol.~0, no.~0, pp. 1--13, 2024.

\bibitem{zheng2024fast}
C.~Zheng, W.~Xu, Z.~Zou, T.~Hua, C.~Yuan, D.~He, B.~Zhou, Z.~Liu, J.~Lin, F.~Zhu \emph{et~al.}, ``Fast-livo2: Fast, direct lidar-inertial-visual odometry,'' \emph{IEEE Transactions on Robotics}, 2024.

\bibitem{nguyen2021viral}
T.-M. Nguyen, M.~Cao, S.~Yuan, Y.~Lyu, T.~H. Nguyen, and L.~Xie, ``Viral-fusion: A visual-inertial-ranging-lidar sensor fusion approach,'' \emph{IEEE Transactions on Robotics}, vol.~38, no.~2, pp. 958--977, 2021.

\bibitem{nguyen2022ntu}
T.-M. Nguyen, S.~Yuan, M.~Cao, Y.~Lyu, T.~H. Nguyen, and L.~Xie, ``Ntu viral: A visual-inertial-ranging-lidar dataset, from an aerial vehicle viewpoint,'' \emph{The International Journal of Robotics Research}, vol.~41, no.~3, pp. 270--280, 2022.

\bibitem{nguyen2024mcd}
T.-M. Nguyen, S.~Yuan, T.~H. Nguyen, P.~Yin, H.~Cao, L.~Xie, M.~Wozniak, P.~Jensfelt, M.~Thiel, J.~Ziegenbein \emph{et~al.}, ``Mcd: Diverse large-scale multi-campus dataset for robot perception,'' in \emph{Proceedings of the IEEE/CVF Conference on Computer Vision and Pattern Recognition}, 2024, pp. 22\,304--22\,313.

\bibitem{xu2023airvo}
K.~Xu, Y.~Hao, S.~Yuan, C.~Wang, and L.~Xie, ``Airvo: An illumination-robust point-line visual odometry,'' in \emph{2023 IEEE/RSJ International Conference on Intelligent Robots and Systems (IROS)}.\hskip 1em plus 0.5em minus 0.4em\relax IEEE, 2023, pp. 3429--3436.

\bibitem{pfreundschuh2024coin}
P.~Pfreundschuh, H.~Oleynikova, C.~Cadena, R.~Siegwart, and O.~Andersson, ``Coin-lio: Complementary intensity-augmented lidar inertial odometry,'' in \emph{2024 IEEE International Conference on Robotics and Automation (ICRA)}.\hskip 1em plus 0.5em minus 0.4em\relax IEEE, 2024, pp. 1730--1737.

\bibitem{xu2025airslam}
K.~Xu, Y.~Hao, S.~Yuan, C.~Wang, and L.~Xie, ``Airslam: An efficient and illumination-robust point-line visual slam system,'' \emph{IEEE Transactions on Robotics}, 2025.

\bibitem{ribeiro2006soi}
A.~Ribeiro, G.~B. Giannakis, and S.~I. Roumeliotis, ``Soi-kf: Distributed kalman filtering with low-cost communications using the sign of innovations,'' \emph{IEEE Transactions on signal processing}, vol.~54, no.~12, pp. 4782--4795, 2006.

\bibitem{trawny2009cooperative}
N.~Trawny, S.~I. Roumeliotis, and G.~B. Giannakis, ``Cooperative multi-robot localization under communication constraints,'' in \emph{2009 IEEE international conference on robotics and automation}.\hskip 1em plus 0.5em minus 0.4em\relax IEEE, 2009, pp. 4394--4400.

\bibitem{msechu2008decentralized}
E.~J. Msechu, S.~I. Roumeliotis, A.~Ribeiro, and G.~B. Giannakis, ``Decentralized quantized kalman filtering with scalable communication cost,'' \emph{IEEE Transactions on Signal Processing}, vol.~56, no.~8, pp. 3727--3741, 2008.

\bibitem{schnabel2006octree}
R.~Schnabel and R.~Klein, ``Octree-based point-cloud compression.'' \emph{PBG@ SIGGRAPH}, vol.~3, no.~3, 2006.

\bibitem{draco_github}
F.~Galligan, M.~Hemmer, O.~Stava, F.~Zhang, and J.~Brettle, ``Google/draco: a library for compressing and decompressing 3d geometric meshes and point clouds,'' 2018.

\bibitem{graziosi2020overview}
D.~Graziosi, O.~Nakagami, S.~Kuma, A.~Zaghetto, T.~Suzuki, and A.~Tabatabai, ``An overview of ongoing point cloud compression standardization activities: Video-based (v-pcc) and geometry-based (g-pcc),'' \emph{APSIPA Transactions on Signal and Information Processing}, vol.~9, p. e13, 2020.

\bibitem{wang2022r}
S.~Wang, J.~Jiao, P.~Cai, and L.~Wang, ``R-pcc: A baseline for range image-based point cloud compression,'' in \emph{2022 International Conference on Robotics and Automation (ICRA)}.\hskip 1em plus 0.5em minus 0.4em\relax IEEE, 2022, pp. 10\,055--10\,061.

\bibitem{wang2022sparse}
J.~Wang, D.~Ding, Z.~Li, X.~Feng, C.~Cao, and Z.~Ma, ``Sparse tensor-based multiscale representation for point cloud geometry compression,'' \emph{IEEE Transactions on Pattern Analysis and Machine Intelligence}, vol.~45, no.~7, pp. 9055--9071, 2022.

\bibitem{jiang2024live}
D.~Jiang, B.~Liu, J.~Wang, A.~Hu, Y.~Zhao, M.~Bao, Z.~Fan, Z.~Shen, K.~Wang, and C.~Wang, ``Live demonstration: A reconfigurable, energy-efficient and high-frame-rate ekf-slam accelerator based soc design for autonomous mobile robot applications,'' in \emph{2024 IEEE International Symposium on Circuits and Systems (ISCAS)}.\hskip 1em plus 0.5em minus 0.4em\relax IEEE, 2024, pp. 1--1.

\bibitem{peng2024quantized}
Y.~Peng, C.~Chen, and G.~Huang, ``Quantized visual-inertial odometry,'' in \emph{2024 IEEE International Conference on Robotics and Automation (ICRA)}.\hskip 1em plus 0.5em minus 0.4em\relax IEEE, 2024, pp. 17\,954--17\,960.

\bibitem{huai2024consistent}
Z.~Huai and G.~Huang, ``A consistent parallel estimation framework for visual-inertial slam,'' \emph{IEEE Transactions on Robotics}, 2024.

\bibitem{cai2025bev}
H.~Cai, S.~Yuan, X.~Li, J.~Guo, and J.~Liu, ``Bev-lio (lc): Bev image assisted lidar-inertial odometry with loop closure,'' \emph{arXiv preprint arXiv:2502.19242}, 2025.

\bibitem{li2024ua}
J.~Li, X.~Xu, J.~Liu, K.~Cao, S.~Yuan, and L.~Xie, ``Ua-mpc: Uncertainty-aware model predictive control for motorized lidar odometry,'' \emph{IEEE Robotics and Automation Letters}, 2025.

\bibitem{li2025limo}
J.~Li, Z.~Liu, X.~Xu, J.~Liu, S.~Yuan, and L.~Xie, ``Limo-calib: On-site fast lidar-motor calibration for quadruped robot-based panoramic 3d sensing system,'' \emph{arXiv preprint arXiv:2502.12655}, 2025.

\bibitem{ji2024sgba}
X.~Ji, S.~Yuan, J.~Li, P.~Yin, H.~Cao, and L.~Xie, ``Sgba: Semantic gaussian mixture model-based lidar bundle adjustment,'' \emph{IEEE Robotics and Automation Letters}, vol.~9, no.~12, pp. 10\,922--10\,929, 2024.

\bibitem{cao2025cooperative}
M.~Cao, T.-M. Nguyen, S.~Yuan, A.~Anastasiou, A.~Zacharia, S.~Papaioannou, P.~Kolios, C.~G. Panayiotou, M.~M. Polycarpou, X.~Xu \emph{et~al.}, ``Cooperative aerial robot inspection challenge: A benchmark for heterogeneous multi-uav planning and lessons learned,'' \emph{arXiv preprint arXiv:2501.06566}, 2025.

\bibitem{liang2025unsupervised}
H.~Liang, Y.~Yang, J.~Hu, J.~Yang, F.~Liu, and S.~Yuan, ``Unsupervised uav 3d trajectories estimation with sparse point clouds,'' in \emph{Proceedings of the IEEE International Conference on Acoustics, Speech, and Signal Processing (ICASSP)}.\hskip 1em plus 0.5em minus 0.4em\relax IEEE, 2025.

\bibitem{xu2024selective}
J.~Xu, G.~Huang, W.~Yu, X.~Zhang, L.~Zhao, R.~Li, S.~Yuan, and L.~Xie, ``Selective kalman filter: When and how to fuse multi-sensor information to overcome degeneracy in slam,'' in \emph{arXiv preprint arXiv:2412.17235}, 2024.

\bibitem{deng2024compact}
T.~Deng, Y.~Chen, L.~Zhang, J.~Yang, S.~Yuan, J.~Liu, D.~Wang, H.~Wang, and W.~Chen, ``Compact 3d gaussian splatting for dense visual slam,'' in \emph{arXiv preprint arXiv:2403.11247}, 2024.

\bibitem{Nguyen2024GPTR}
T.-M. Nguyen, Z.~Cao, K.~Li, S.~Yuan, and L.~Xie, ``Gptr: Gaussian process trajectory representation for continuous-time motion estimation,'' in \emph{arXiv preprint arXiv:2410.22931}, 2024.

\bibitem{bai2024collaborative}
R.~Bai, S.~Yuan, H.~Guo, P.~Yin, W.-Y. Yau, and L.~Xie, ``Collaborative graph exploration with reduced pose-slam uncertainty via submodular optimization,'' in \emph{Proceedings of the 2024 IEEE/RSJ International Conference on Intelligent Robots and Systems (IROS)}.\hskip 1em plus 0.5em minus 0.4em\relax IEEE, 2024.

\bibitem{yu2024i2ekf}
W.~Yu, J.~Xu, C.~Zhao, L.~Zhao, T.-M. Nguyen, S.~Yuan, M.~Bai, and L.~Xie, ``I2ekf-lo: A dual-iteration extended kalman filter based lidar odometry,'' in \emph{2024 IEEE/RSJ International Conference on Intelligent Robots and Systems (IROS)}, 2024, pp. 10\,453--10\,460.

\bibitem{ji2022robust}
T.~Ji, S.~Yuan, and L.~Xie, ``Robust rgb-d slam in dynamic environments for autonomous vehicles,'' in \emph{2022 17th International Conference on Control, Automation, Robotics and Vision (ICARCV)}.\hskip 1em plus 0.5em minus 0.4em\relax IEEE, 2022, pp. 665--671.

\bibitem{nguyen2021liro}
T.-M. Nguyen, M.~Cao, S.~Yuan, Y.~Lyu, T.~H. Nguyen, and L.~Xie, ``Liro: Tightly coupled lidar-inertia-ranging odometry,'' in \emph{2021 IEEE international conference on robotics and automation (ICRA)}.\hskip 1em plus 0.5em minus 0.4em\relax IEEE, 2021, pp. 14\,484--14\,490.

\bibitem{zhang2014loam}
J.~Zhang, S.~Singh \emph{et~al.}, ``Loam: Lidar odometry and mapping in real-time.'' in \emph{Robotics: Science and systems}, vol.~2, no.~9.\hskip 1em plus 0.5em minus 0.4em\relax Berkeley, CA, 2014, pp. 1--9.

\bibitem{shan2018lego}
T.~Shan and B.~Englot, ``Lego-loam: Lightweight and ground-optimized lidar odometry and mapping on variable terrain,'' in \emph{2018 IEEE/RSJ International Conference on Intelligent Robots and Systems (IROS)}.\hskip 1em plus 0.5em minus 0.4em\relax IEEE, 2018, pp. 4758--4765.

\bibitem{liu2024voxel}
Z.~Liu, H.~Li, C.~Yuan, X.~Liu, J.~Lin, R.~Li, C.~Zheng, B.~Zhou, W.~Liu, and F.~Zhang, ``Voxel-slam: A complete, accurate, and versatile lidar-inertial slam system,'' \emph{arXiv preprint arXiv:2410.08935}, 2024.

\bibitem{yuan2024btc}
C.~Yuan, J.~Lin, Z.~Liu, H.~Wei, X.~Hong, and F.~Zhang, ``Btc: A binary and triangle combined descriptor for 3-d place recognition,'' \emph{IEEE Transactions on Robotics}, vol.~40, pp. 1580--1599, 2024.

\bibitem{nerurkar2013communication}
E.~D. Nerurkar and S.~I. Roumeliotis, ``A communication-bandwidth-aware hybrid estimation framework for multi-robot cooperative localization,'' in \emph{2013 IEEE/RSJ International Conference on Intelligent Robots and Systems}.\hskip 1em plus 0.5em minus 0.4em\relax IEEE, 2013, pp. 1418--1425.

\end{thebibliography}

\end{document}